\documentclass{article}

     \PassOptionsToPackage{numbers, compress}{natbib}

\usepackage[preprint]{neurips_2026}

\usepackage[utf8]{inputenc} 
\usepackage[T1]{fontenc}    
\usepackage{hyperref}       
\usepackage{url}            
\usepackage{booktabs}       
\usepackage{amsfonts}       
\usepackage{nicefrac}       
\usepackage{microtype}      
\usepackage{xcolor}         

\usepackage{amsmath}
\usepackage{booktabs}
\usepackage{multirow}
\usepackage{subcaption}
\usepackage{graphicx}
\usepackage{wrapfig}
\usepackage{etoc}
\usepackage{titletoc}
\usepackage{amssymb} 

\title{Evolution-Aware MSA Reasoning for Subsampling via Factor Graphs}

\author{%
  {\bf Zhangzhi Xiong}$^{1,*}$\quad
  {\bf Minzhang Li}$^{1,*}$\quad
  {\bf Haotian Yu}$^{1,*}$
  \\
  {\bf Sixian Shen}$^{1}$\quad
  {\bf Kexin Zhang}$^{1}$\quad
  {\bf Mingrui Li}$^{1}$
  \\
  {\bf Jie Zheng}$^{1,\dagger}$\quad
  {\bf Kewei Tu}$^{1,\dagger}$\quad
  {\bf Jingyi Yu}$^{1,\dagger}$
  \\[0.2em]
  $^{1}$School of Information Science and Technology, ShanghaiTech University
  \\[0.2em]
  \texttt{xiongzhzh2023@shanghaitech.edu.cn}
  \\[0.2em]
  $^{*}$Equal contribution.\quad
  $^{\dagger}$Corresponding authors.
}

\begin{document}

\maketitle

\begin{abstract}

Multiple Sequence Alignments (MSAs) provide protein language models with explicit evolutionary context, but their large depth makes subsampling unavoidable under limited token budgets. Existing strategies, including random selection, identity-based filtering, and diversity-driven sampling, are effective heuristics, yet provide limited control over the evolutionary signals retained in the subset. In this work, we recast MSA subsampling as an explicit optimization problem, where key evolutionary measures, including query identity and diversity, are treated as controllable objectives. Building on this view, we introduce AP-REASONER, an Affinity-Propagation-based factor-graph approach. With evolution-aware unary factors, exemplar-consistency factors, and two control knobs, AP-REASONER performs factor-graph reasoning through message passing to infer a fixed-budget MSA subset. Experiments on long-range contact prediction and conformational ensemble prediction show that AP-REASONER outperforms baseline subsamplers on structure-sensitive downstream tasks and enables controllable recovery of alternative protein conformations. These results highlight the value of modeling MSA subsampling as a controllable optimization problem, where factor-graph reasoning offers an effective alternative to heuristic selection.
\end{abstract}

\section{Introduction} \label{sec:introduction}

Evolution has profoundly shaped and diversified the biological language of proteins over eons~\citep{anfinsen1973principles,chothia1986relation}. Across this vast temporal scale, the sequence identity preserved by shared ancestry couples with the diversity introduced by stochastic mutations, insertions, and deletions, collectively defining a constrained and richly populated sequence space~\citep{morcos2011direct,marks2011protein,hopf2017mutation}. This sequence space is, in essence, a functional landscape that deeply encodes the physics of protein folding and biological constraints~\citep{dill2012protein,jumper2021highly,lin2023evolutionary}. Consequently, precisely and rationally modeling it has emerged as the pivotal key for diverse downstream applications, including contact and structure prediction~\citep{senior2020improved,jumper2021highly,baek2021accurate,abramson2024accurate,wang2017accurate}, dynamic conformational analysis~\citep{del2022sampling,wayment2024predicting}, and drug discovery~\citep{abramson2024accurate,frazer2021disease,evans2021protein}.

The transition from evolutionary theory to computational modeling has birthed two divergent paradigms for ``presenting'' evolutionary signals to protein language models (PLMs)~\citep{bert,tape}. Single-sequence models~\citep{rives2021biological,lin2023evolutionary,heinzinger2019modeling,elnaggar2021prottrans,notin2022tranception,madani2023large} relegate these signals to an implicit prior; by training on vast unaligned databases, they leverage generative objectives to internalize conservation patterns and structural constraints directly within their neural parameters~\citep{vig2020bertology,DBLP:conf/iclr/RaoMSOR21}. In contrast, MSA-based models~\citep{rao2021msa,jumper2021highly,abramson2024accurate,hong2022prot,yang2020improved} adopt a more explicit approach, employing Multiple Sequence Alignments (MSAs)~\citep{steinegger2019hh,mirdita2022colabfold,steinegger2017mmseqs2} as a discrete representation of the evolutionary manifold to provide a stronger inductive bias~\citep{rao2021msa,ahdritz2023openproteinsettrainingdatastructural}. However, this explicit dependency is stifled by a severe computational trade-off: due to the quadratic complexity of attention mechanisms, such models can only process a narrow context of a few hundred sequences~\citep{rao2021msa,jumper2021highly}---a mere fraction of the rich evolutionary information available in genomic databases~\citep{steinegger2017mmseqs2,mirdita2022colabfold}.

\begin{figure}[t] \label{fig:intro}
  \centering
\includegraphics[width=1.0\linewidth]{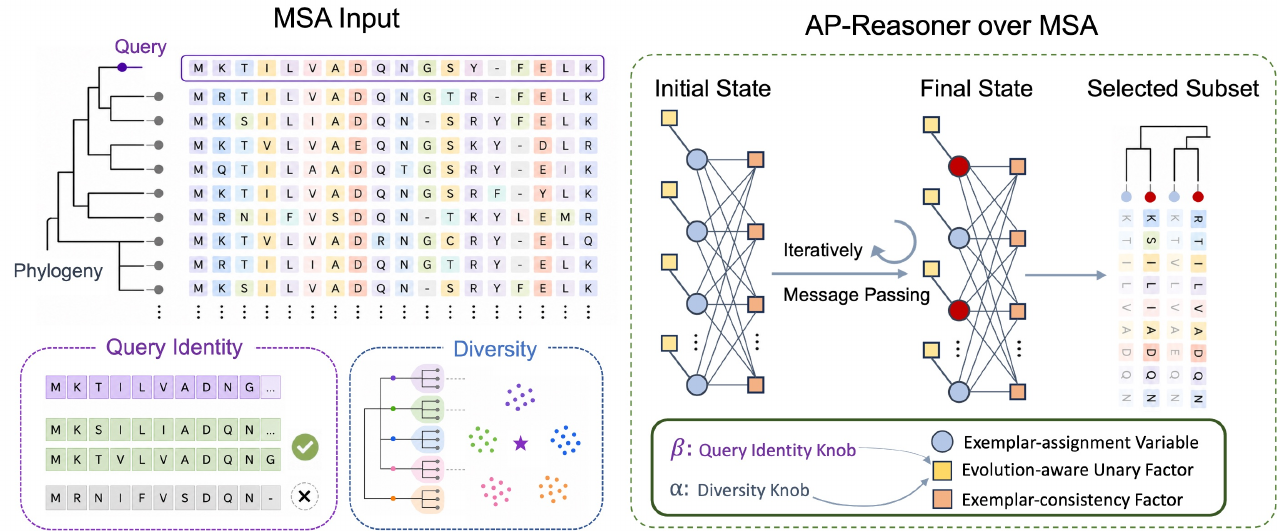}%
    \caption{The left part illustrates two main MSA indicators for evolution: query identity and diversity. The right part shows the structure of \ap{}, where red nodes are chosen as the MSA subset.}
    \vspace{-10pt}
\end{figure}

To handle the vast repertoire of homologs within constrained computational budgets, the prevailing paradigm, exemplified by AlphaFold2~\citep{jumper2021highly} and MSA Transformer~\citep{rao2021msa}, relies on heuristic subsampling to fit sequences into narrow attention windows~\citep{stein2022speach_af,wayment2024predicting}. Whether through identity-based filtering via HHfilter~\citep{steinegger2019hh}, clustering via MMseqs2~\citep{steinegger2017mmseqs2} or diversity-driven samplers such as divMax and divMin~\citep{rao2021msa,wayment2024predicting}, these methods typically reduce MSA subsampling to threshold-based filtering, clustering, or single-criterion selection under pairwise sequence-alignment metric~\citep{petti2023end}. While effective for removing near-duplicate sequences or biasing samples toward a desired diversity regime, they primarily operate on local similarity relations and do not explicitly model trade-offs w.r.t. diversity and query identities. This makes the explicit balancing of these two complementary aspects---query identity~\citep{chothia1986relation,steinegger2019hh} and diversity~\citep{rao2021msa,wayment2024predicting}--- a natural and worthwhile direction for MSA subsampling. 

Building upon this insight, we recast MSA subsampling as an explicit optimization problem rather than a heuristic sampling procedure, where biologically meaningful evolutionary measures such as diversity and query identity are elevated from heuristics to explicit objectives. This formulation allows subset assignments to be coordinated under carefully designed selection objectives. Leveraging this perspective, we propose \ap{} for MSA subsampling with Affinity Propagation (AP)~\citep{frey2007clustering}. Utilizing biologically designed unary factors and exemplar-consistency factors, AP's exemplar-based message-passing reasoning provides a principled way to calibrate preferences within both sequence-diversity and query-identity aspects during MSA subsampling. To make this balance explicitly controllable, we further design two control knobs within the AP unary factor to regulate these two aspects respectively.

The contributions of this work are threefold. First, to the best of our knowledge, we are the first to recast MSA subsampling from random or heuristic selection to an explicit optimization problem. Second, we introduce \ap{}, an Affinity-Propagation-based factor-graph approach that encodes the subsampling objective into factors and solves the resulting optimization problem through message-passing reasoning, with two control knobs in the unary factor to explicitly modulate sequence diversity and query identity. Third, evaluations on long-range contact prediction and conformational ensemble prediction show that \ap{} consistently outperforms baseline subsamplers, highlighting MSA factor-graph optimization reasoning as an important paradigm for downstream performance and validating the two control knobs as an effective design.

\section{Related Work}
\label{sec:related-work}

\paragraph{MSA subsampling.}
MSA-based PLMs reach state-of-the-art on structure-related tasks~\citep{jumper2021highly, abramson2024accurate}, but the full MSA is rarely tractable: a query against UniRef or BFD~\citep{mirdita2022colabfold} routinely returns tens of thousands of homologs while attention-based PLMs can only condition on a small fraction of them~\citep{rao2021msa, jumper2021highly}, making subsampling a constitutive choice rather than an option. Existing strategies are simple heuristics applied only at inference on top of a pretrained backbone: random selection~\citep{del2022sampling}; \texttt{HHfilter}~\citep{steinegger2019hh} by identity threshold; AFcluster~\citep{wayment2024predicting} via DBSCAN for multiple conformation prediction, where biologically functional states can be recovered from MSAs; the greedy \emph{divmin} and \emph{divmax}~\citep{rao2021msa, hong2022prot} pushing each pick toward minimum or maximum diversity. How a subsampling strategy preserves an MSA's evolutionary structure, and how that choice propagates to training, remains largely unstudied. We therefore propose a factor-graph approach that imposes evolutionary structure by reasoning over the candidate homolog pool rather than filtering it.

\paragraph{Protein language models.}
Protein language models (PLMs) apply NLP-style sequence modeling to amino acids, learning residue-level representations that encode structural, functional, and evolutionary information~\citep{rives2021biological,heinzinger2019modeling,vig2020bertology}. Long-range residue contact prediction has become a key benchmark for evolutionary understanding, since sequence-distant but spatially close residue pairs can be recovered from co-evolutionary signal~\citep{tape}. Single-sequence PLMs, including ESM-1b~\citep{rives2021biological}, SeqVec~\citep{heinzinger2019modeling}, and Tranception~\citep{notin2022tranception}, learn such signals implicitly and scale efficiently, but this implicit treatment often requires larger model scale~\citep{lin2023evolutionary}, can be less reliable for long-range dependencies and rare evolutionary variants without external context~\citep{jumper2021highly}, and provides limited direct control for downstream settings such as conformational ensemble generation~\citep{del2022sampling,wayment2024predicting,SPEACHAF,Lee2025CFRandom,MonteiroDaSilva2024SubsampledAlphaFold2}. In contrast, MSA-based PLMs such as MSA Transformer~\citep{rao2021msa} treat the MSA as a two-dimensional sequence-by-position input, applying row and column attention to externalize family-level evolutionary signal in the alignment itself; this design is parameter-efficient for long-range contact prediction~\citep{rao2021msa}. Related AlphaFold-family architectures, including Evoformer~\citep{jumper2021highly} and Pairformer~\citep{abramson2024accurate}, further demonstrate the importance of MSA- or pair-conditioned representations for structure modeling.

\paragraph{Factor graphs for joint reasoning.}
Factor graphs are a foundational probabilistic-modeling and optimization framework, going back to Pearl's belief propagation~\citep{pearl2014probabilistic} and the sum-product algorithm of Kschischang, Frey, and Loeliger~\citep{kschischang2001factor}, with a modern variational view from Wainwright and Jordan~\citep{wainwright2008graphical}. They have driven progress across coding theory, statistics, computer vision~\citep{krahenbuhl2011efficient}, and natural language processing~\citep{huang2015bidirectional}. The framework spans Conditional Random Fields~\citep{lafferty2001conditional}, undirected Markov random fields, directed Bayesian networks and HMMs, etc, all formulating inference as an energy-minimization or MAP optimization problem solved by message passing. Affinity propagation~\citep{frey2007clustering} is a classical method that jointly optimizes the selection of representatives from a candidate set, with unary and exemplar-consistency factors giving clean control over query identity and redundancy (rationale in Section~\ref{sec:method}). At inference, reasoning jointly over the pool, AP passes responsibility and availability messages on the factor graph until a stable set of exemplars emerges.

\section{Method}
\label{sec:method}
\subsection{MSA Subsampling: Background}

Multiple Sequence Alignments (MSAs) provide MSA-based protein language models with rich evolutionary context~\citep{rao2021msa, jumper2021highly}. Let $\mathbf{X} = (x_0, x_1, \ldots, x_{M-1}) \in (\mathcal{A} \cup \{-\})^{M \times L}$ denote an input MSA, where $x_0$ is the primary query sequence, each aligned row $x_i$ has length $L$, $\mathcal{A}$ denotes the vocabulary of amino acids and $-$ represents an alignment gap. In practice, the number of sequences in an MSA, $M$, can be overwhelmingly large, often reaching thousands or more. 

This scale presents a computational bottleneck for the training of protein language models (PLMs). MSA-based PLMs such as MSA Transformer~\citep{rao2021msa} process an input alignment as a 2D grid with $R$ rows and $L$ columns, and their axial self-attention scales as $\mathcal{O}(R L^2 + L R^2)$ per layer. Consequently, given a fixed token budget $B$, length $L$ induces a maximum admissible budget of rows:
$$
N_B(L) = \min\!\left(\left\lfloor \frac{B}{L} \right\rfloor, M\right)
$$
When $M > N_B(L)$, MSA must be subsampled to $N_B(L)$ while strictly retaining the query sequence.

Existing subsamplers provide simple mechanisms for fitting large MSAs into fixed token budgets. 
\texttt{Random} subsampling is cheap and budget-compliant, but it offers no explicit control over the evolutionary composition of the selected set. 
Heuristic alternatives such as \texttt{HHfilter}~\citep{steinegger2019hh} and \texttt{divMax}~\citep{rao2021msa} introduce more structured preferences, targeting dereplication or diversity promotion through query identity thresholds or greedy sequential choices. 
These methods are therefore best viewed as effective engineering heuristics for enforcing specific sampling biases, rather than as a general mechanism for coordinating multiple evolutionary preferences. In contrast, we recast MSA subsampling from heuristic sampling into an explicit optimization problem.
This allows us to define the desired evolutionary measures, including query identity and diversity, as optimization objectives and obtain MSA subsets whose composition is more consistent with the modeled preferences.


\subsection{\ap{}}
\label{method:APReasoner}


We provide a factor-graph formulation of the MSA subsampling problem and hence cast MSA subsampling as a maximum a posteriori problem on the factor graph. Specifically, our formulation follows that of Affinity Propagation (AP) \citep{frey2007clustering}, an exemplar-based clustering model, and injects MSA evolution-relevant preferences such as sequence diversity and query identity into the factors. We refer to our AP-based subsampling method as \ap{}.


For each row $i$, we introduce a discrete variable
$s_i \in \{0, \ldots, M-1\}$, where $s_i = k$ means that row $i$ is represented by row $k$.
A row $k$ is an exemplar if and only if $s_k = k$.
We retain the original factor-graph topology of affinity propagation, where each row-wise assignment variable is associated with a unary likelihood factor
$L_i(s_i)=\exp(S_{i,s_i})$, with $S\in\mathbb{R}^{M\times M}$ denoting the similarity matrix.
In addition, the model includes the standard exemplar-consistency factors $f_k$, which impose the constraint that if any row selects row $k$ as its exemplar, then row $k$ must also select itself.
As we adopt AP's standard practical implementation of $f_k$, we defer its detailed definition to Appendix~\ref{si:fk}.

Conceptually, the unary factor $L_i(s_i)=\exp(S_{i,s_i})$ measures how plausible it is for row $i$ to be represented by candidate exemplar $s_i$; within it, the off-diagonal score $S_{ik}\,(i\neq k)$ quantifies how well row $k$ represents row $i$, whereas the diagonal score $S_{kk}$ encodes how suitable row $k$ is to declare itself an exemplar. This motivates our separate definitions of off-diagonal similarities ($S_{ik}\,(i\neq k)$) and self-preferences ($S_{kk}$) in the matrix $S$, through which we explicitly incorporate two evolutionarily important measures: sequence diversity and query identity.

\begin{wrapfigure}[8]{r}{0.54\textwidth}
  \centering
  \includegraphics[width=0.53\textwidth]{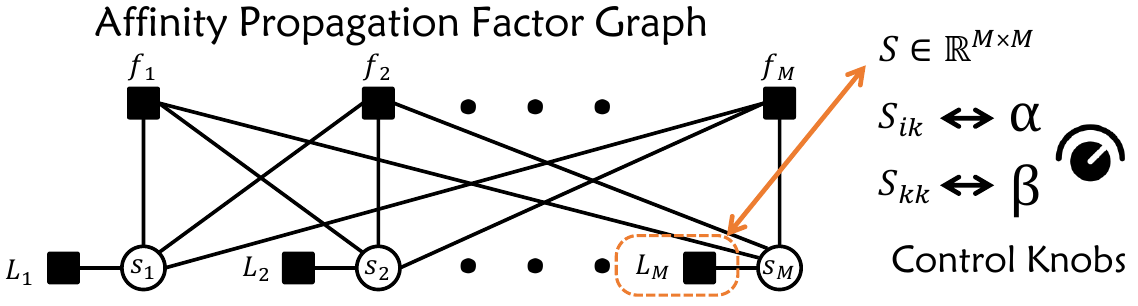}
  \caption{AP factor graph and control knob. }
  \label{fig:intro}
\end{wrapfigure}

We adopt normalized gap-aware Hamming distance as an alignment-native proxy for sequence redundancy and divergence, in line with the residue-disagreement measures widely used in evolutionary-coupling analysis~\citep{morcos2011direct}. It measures direct residue disagreement at homologous positions and avoids auxiliary model assumptions. While bypassing complex evolutionary discrepancies, it sufficiently preserves the coarse structural neighborhood essential for subsampling: effectively isolating locally redundant sequences from those providing complementary variation. Hamming distance can be formulated as:
\begin{equation}
d_H^{\mathrm{gap}}(i,k)
=
\left|
\left\{
\ell \in \{1,\ldots,L\}
:
x_{i\ell} \neq -,\;
x_{k\ell} \neq -,\;
x_{i\ell} \neq x_{k\ell}
\right\}
\right|,
\end{equation}
which counts mismatched columns where both rows are non-gap. We then define off-diagonal entries
\begin{equation}
\label{eq:alpha}
S_{ik}
=
-\left(\frac{d_H^{\mathrm{gap}}(i,k)}{L}\right)^{\alpha},
\qquad i \neq k,
\end{equation}
where $\alpha$ controls distance curvature. 

For diagonal entries, we define self-preferences as
\begin{equation}
\label{eq:beta}
S_{kk} = p + \beta \frac{d_H^{\mathrm{gap}}(k,0)}{L},
\qquad k \neq 0,
\end{equation}
where $p$ is a global preference offset and a row-dependent query-distance prior is injected into exemplar selection with coefficient $\beta$. 

Note that we normalize the Hamming Distance count by the aligned length $L$ to obtain a length-normalized dissimilarity. Furthermore, following the subsampling setting in MSA Transformer~\citep{rao2021msa}, the query row is pinned by a query-anchor preference $S_{00} = p_q$, with $p_q$ chosen sufficiently large so that the query is always retained as an exemplar.

To perform global optimization on the factor graph, our implementation follows the log-domain max-product form of AP reasoning popularized by~\citet{frey2007clustering}. More details can be referred to Appendix~\ref{si:logreasoning}.

\paragraph{Two Control Knobs.} The factor formulations above provide two control knobs, $\alpha$ and $\beta$, over \ap{} for MSA subsampling. Specifically, the exponent $\alpha \in \mathbb{R}^+$ acts as a control knob for global diversity. Mathematically, a larger $\alpha$ flattens the penalty for small fractional distances, making the model highly tolerant of minor local mutations. This effectively collapses local redundancies, forcing the selected exemplars to be more globally distributed, thereby yielding a subset with higher diversity. In complement, the coefficient $\beta \in \mathbb{R}$ explicitly regulates the exemplar-to-query identity by injecting a query-distance prior into the diagonal self-preferences. A larger positive $\beta$ strongly incentivizes the selection of exemplars that are distant from the query, whereas a negative $\beta$ biases the reasoning engine to favor similar exemplars.

\paragraph{Fixed-cardinality Wrapper.}
AP controls the exemplar count implicitly via the self-preference $p$. In order to conform to the exact row budget $N_B(L)$, we wrap AP with a fixed-cardinality procedure: bisection over $p$ drives the exemplar count close to $N_B(L)$, then top-$N$ truncation or padding by the diagonal belief $a(k,k)+r(k,k)$ yields an exact-$N$ subset. The implementation detail is described in Appendix ~\ref{si:wrapper}.

\section{Experiments}
We conduct experiments to evaluate \ap{} from four complementary perspectives: protein language pretraining, long-range contact prediction, protein multiple conformations prediction, and targeted empirical validation of its control knobs $\alpha$ and $\beta$.

\subsection{Protein Language Model Training}
\label{exp:plm_train}
We conduct unsupervised protein language model pretraining under the standard masked language modeling (MLM) objective, in which a fraction of input MSA tokens is stochastically masked and the model is trained to recover the original amino acids from the remaining sequences context. More training details are described in Appendix~\ref{si:mlm}. 

\begin{table}[t]
\centering
\small
\setlength{\tabcolsep}{5pt}
\renewcommand{\arraystretch}{1.05}
\caption{MLM evaluation on a random-subsampled test set. Columns denote checkpoints trained with different MSA subsampling strategies. All models are evaluated on the same random-sampled test set of 10K MSAs.}
\label{tab:mlm-random-test}
\begin{tabular}{lcccccc}
\toprule
Metric & random & divMin & divMax & AP & K-medians & HHfilter \\
\midrule
Loss             & 1.0932 & 1.2038 & 1.1298 & 1.1000 & 1.1029 & 1.1139 \\
PPL              & 2.98 & 3.33 & 3.10 & 3.00 & 3.01 & 3.05 \\
Denoising Acc    & 67.5\% & 65.3\% & 66.7\% & 67.3\% & 67.3\% & 67.0\% \\
\bottomrule
\end{tabular}
\vspace{-10pt}
\end{table}

As presented in Table~\ref{tab:mlm-random-test}, on the random-subsampled MLM test set, the random-trained checkpoint performs best overall, while AP and K-medians remain competitive and clearly outperform divMin and divMax. These results suggest that, in terms of MLM performance, random subsampling is already a  strong and fair baseline, while both K-medians and AP remain fully comparable alternatives. However, as revealed in TAPE~\citep{tape}, strong performance on the MLM evaluation does not necessarily translate into strong downstream performance.

\subsection{Long Range Contact Prediction}\label{exp:lrcp}

Long-range contact prediction (LRCP) is biologically important because it probes whether a model captures nonlocal residue couplings that are critical to protein three-dimensional structure and folding. We evaluate long range contact prediction following \citet{rao2021msa}. Specifically, we fit a minimal sparse regression head on frozen row-attention features using 20 protein structures from the RCSB~\citep{rcsb} Protein Data Bank, so that the evaluation isolates the quality of the pretrained representations rather than end-to-end supervised adaptation. Contact prediction is assessed using long-range \(\mathrm{C}\beta\mathrm{-}\mathrm{C}\beta\) contacts (< 8\,\AA; \(\mathrm{C}\alpha\) for glycine), and we report Top-\(K\) precision for \(K \in \{L, L/2, L/5\}\), where \(L\) is the target protein length, considering only residue pairs with \(|i-j| \geq 24\). During testing inference, all CASP15~\citep{casp15} MSAs are subsampled to a depth of 256. We evaluate on MSA Transformer checkpoints pretrained with six subsampling strategies mentioned in Section~\ref{exp:plm_train}---random, divMax, HHfilter, divMin, \ap{}, and K-medians---and report CASP15 long-range contact prediction precision on both the full set and the 36-domain informative subset obtained after excluding 9 universal-failure domains. Complete details of the regression head, dataset, test set settings, evaluation metric and informative subset are provided in Appendix~\ref{si:lrcp}.

Table~\ref{tab:divmax-casp15-topl} reports the Top-L long range contact precision, where models trained with \ap{} consistently outperform all models trained with baseline samplers. Notably, AP achieves a substantial margin over K-medians and traditional heuristic HHfilter, with the performance gap widening on the informative subset, underscoring the superior representation quality learned via reasoning-based subsampling. The result tables of Top-L/2 and Top-L/5 are presented in Appendix~\ref{si:lrcp_more}. As shown in Figure~\ref{fig:hump}, different MSA subsampling strategies for PLM training lead to visibly different contact-map quality, with AP generally recovering cleaner long-range contact structure than the weaker baselines.

\begin{table}[t]
\centering
\small
\setlength{\tabcolsep}{6pt}
\renewcommand{\arraystretch}{1.05}
\caption{Top-L long-range contact precision on divMax-sampled CASP15 MSAs. Columns denote training-time samplers (checkpoints).}
\label{tab:divmax-casp15-topl}
\begin{tabular}{lcccccc}
\toprule
Setting & random & divMin & divMax & kmedians & HHfilter & AP \\
\midrule
Full CASP15 ($n=45$)         & 31.5 & 17.2 & 27.7 & 25.9 & 22.3 & \textbf{35.1}\\
Informative subset ($n=36$)  & 38.8 & 21.0 & 34.2 & 31.9 & 27.3 & \textbf{43.3}\\
\bottomrule
\end{tabular}
\vspace{-10pt}
\end{table}

\begin{figure}[tb] 
    \centering 
    \includegraphics[width=\textwidth]{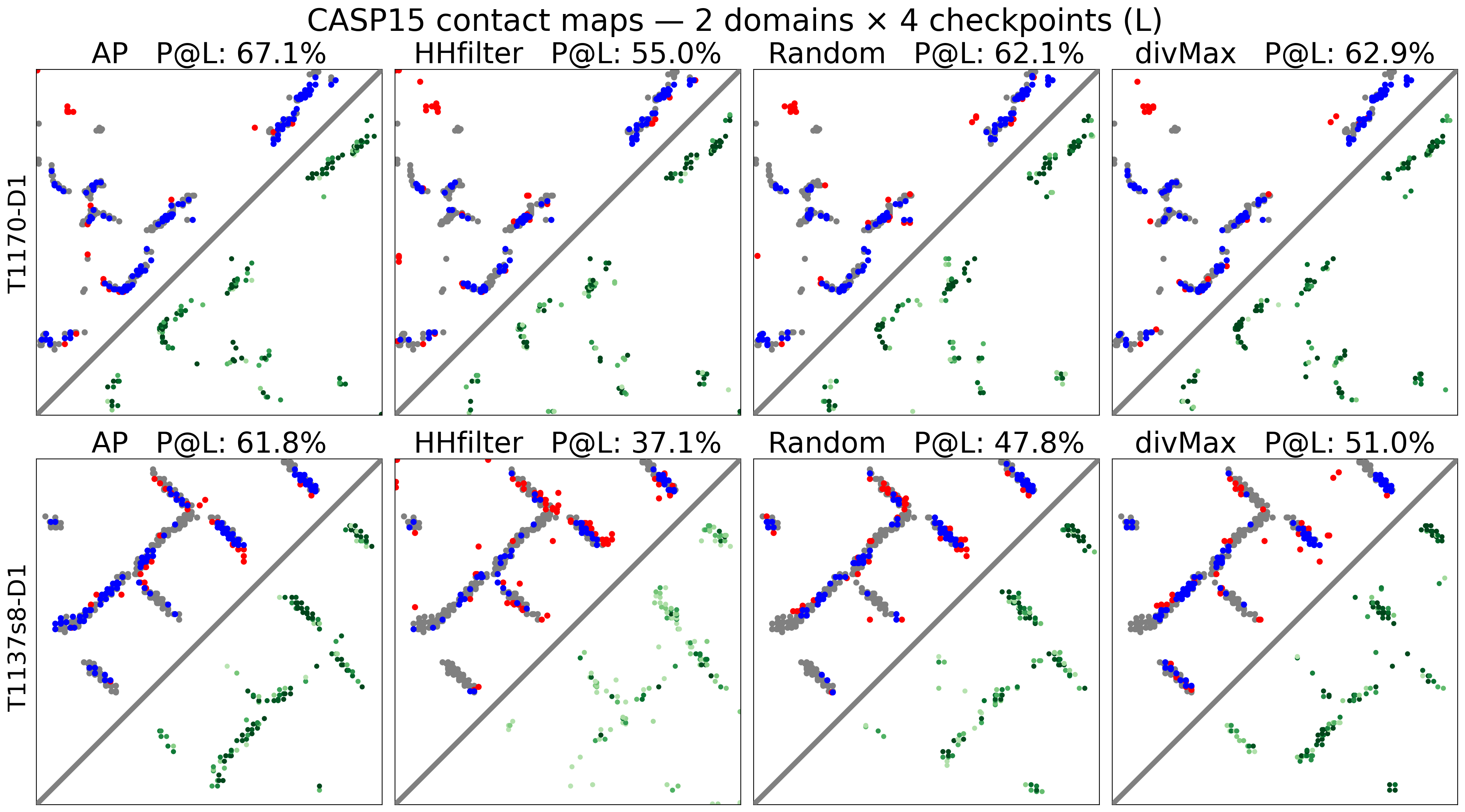} 
    \caption{Contact prediction under different MSA subsampling strategies on CASP15 targets. We compare AP, HHfilter, Random and divMax on two domains, reporting Top-L contact precision P@L above each contact map. Raw contact probabilities are represented by green points below the diagonal, while Top-L predicted contacts are shown above the diagonal; blue, red, and grey points denote true positives, false positives, and ground-truth contacts, respectively.}
    \label{fig:hump}
    \vspace{-15pt}
\end{figure}

Motivated by MSA Transformer, which showed that MSA selection with greater diversity improves long-range contact prediction performance~\cite{rao2021msa}, we asked whether the $\alpha$ in \ap{} can serve as an control knob for diversity. The settings of MSA transformer training and downstream long range contact prediction experiment are consistent with Section~\ref{exp:plm_train} and Section~\ref{exp:lrcp}. We evaluate checkpoints trained with six different MSA subsampling strategies and apply either divMax or \ap{} with varying diversity exponent $\alpha$ to subsample the full CASP15 MSAs at test time.

\begin{table}[h]
\centering
\small
\setlength{\tabcolsep}{4pt}
\renewcommand{\arraystretch}{1.05}
\caption{Top-L/5 Long range contact prediction results on informative CASP15 36-domain subset obtained after excluding 9 universal-failure domains. Rows denote training-time samplers (checkpoints), and columns denote inference-time samplers. We retain AP-based inference with varying diversity exponent $\alpha$, together with the divMax inference baseline.}
\label{tab:topl5_informative_alpha}
\resizebox{\linewidth}{!}{
\begin{tabular}{lcccccc}
\toprule
ckpt & AP ($\alpha=0.4$) & AP ($\alpha=0.7$) & AP ($\alpha=1.0$) & AP ($\alpha=1.3$) & AP ($\alpha=1.6$) & divMax \\
\midrule
random   & 53.5 & 53.3 & 53.5 & 54.3 & 54.7 & 59.1 \\
divMax   & 50.9 & 52.1 & 49.5 & 51.7 & 52.2 & 60.4 \\
HHfilter & 37.5 & 37.1 & 38.3 & 39.3 & 39.8 & 44.7 \\
divMin   & 25.4 & 24.3 & 25.5 & 26.6 & 26.7 & 28.2 \\
ap       & 60.7 & 61.3 & 60.1 & 60.1 & 59.9 & 64.7 \\
K-medians & 46.0 & 46.8 & 47.8 & 50.1 & 50.2 & 43.9 \\
\midrule
col mean & \textbf{45.7} & \textbf{45.8} & \textbf{45.8} & \textbf{47.0} & \textbf{47.3} & \textbf{50.2} \\
\bottomrule
\end{tabular}
}
\vspace{-10pt}
\end{table}

Table~\ref{tab:topl5_informative_alpha} provides evidence under the tight Top-L/5 budget on the informative CASP15 subset. While divMax remains the strongest inference sampler overall, the AP column means increase monotonically with $\alpha$ (45.7, 45.8, 45.8, 47.0, and 47.3 for $\alpha=0.4,0.7,1.0,1.3,1.6$, respectively), indicating that larger $\alpha$ improves LRCP performance on average, consistent with its intended role in promoting more diverse AP-selected subsets. This trend is also visible for several individual checkpoints and for several settings whose corresponding additional cross-sampler results are provided in Appendix~\ref{si:alpha-result}.
\subsection{Protein Multiple Conformations Prediction}
\label{exp:ensemble}

\begin{figure}[t]
  \centering
  \noindent\makebox[\linewidth][c]{%
    \includegraphics[width=\linewidth]{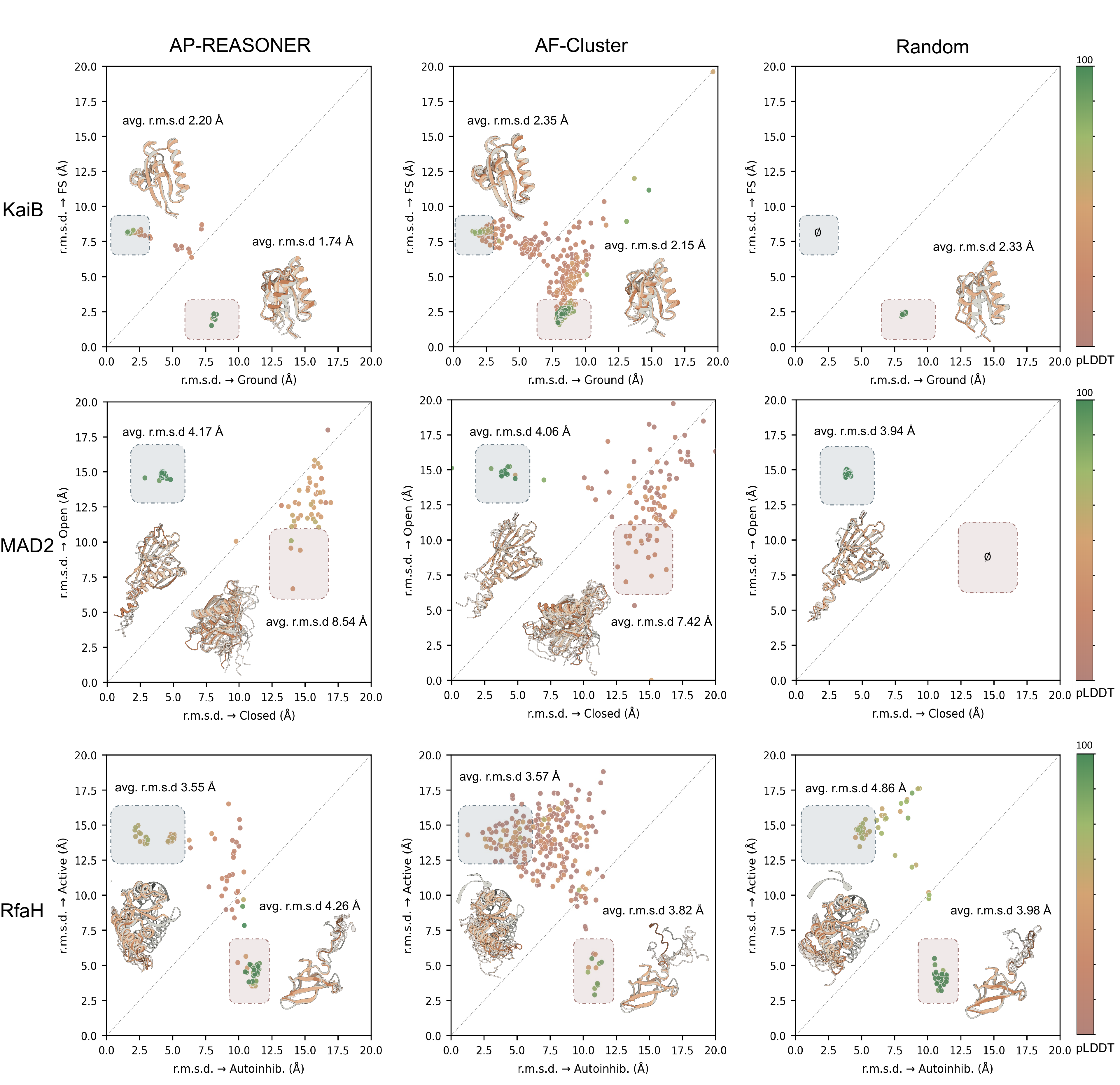}%
  }
  \caption{Comprehensive display for the r.m.s.d. of protein structure predictions (conformations prediction). \ap{} ensembles to high confidence region with much less steps compared to AF-Cluster. The random method may miss one of the high confidence region.}
  \label{fig:knob_alpha}
  \vspace{-15pt}
\end{figure}

Understanding protein function requires not only  predicting contact relationships accurately, but also mapping its full conformational ensemble. While it was shown that heuristically subsampling the input MSA enables AF2 to predict known conformational changes of transporters, such approaches leave the specific evolutionary signals unmodeled and uncontrollable, leading to poor sampling efficiency which causes a massive waste of GPU resources. \ap{}, however, enables the direct deconvolution of state-specific coupling signals. Our following experiments demonstrate that by explicitly steering the sampling process through our tunable control knob, we can predictably recover target conformations that are otherwise lost in random subsampling. This not only proves the controllability of our approach but also highlights its potential in discovering rare functional states that are critical for understanding biological mechanisms. In particular, we evaluate on three well-characterized fold-switching proteins: KaiB, RfaH, and MAD2, possessing at least two experimentally deposited conformational states. 
\paragraph{Results.} As illustrated in Figure \ref{fig:knob_alpha} and Table \ref{tab:pmcp_box_stats}, \ap{} demonstrates a significant advantage in sampling precision and computational efficiency. While heuristic baselines like random subsampling often fail to recover less dominant states, our method consistently populates these regions with high-confidence predictions. This is particularly evident in cases such as the Ground state of KaiB and the Open state of Mad2, where stochastic methods result in empty clusters. 

The superiority of the control knob mechanism is most apparent in the resulting sampling density. For the FS state of KaiB, \ap{} achieves a higher average structural accuracy of 2.20 \text{\AA} RMSD using only 56 targeted samples. In contrast, AF-Cluster requires a brute-force set of 1008 samples to reach a comparable space of 2.35 \text{\AA}. This nearly 20-fold reduction in sampling volume highlights the effectiveness of deconvolving state-specific signals rather than relying on stochastic MSA perturbation. Furthermore, on challenging fold-switching targets like RfaH, \ap{} yields best-in-class recovery for the Autoinhibited state with a best RMSD of 2.15 \text{\AA}. This significantly outperforms the 4.54 \text{\AA} achieved by random subsampling. Figure \ref{fig:knob_alpha} visually confirms these findings by showing that the ensembles generated by \ap{} tightly cluster within the high-confidence regions of the ground-truth boxes. The distributions produced by AF-Cluster and random methods appear more diffused or incomplete. These results prove that \ap{} provides a controllable mechanism for conformation discovery while ensuring that the resulting models remain high-fidelity and biologically relevant.

\subsection{Qualitative Analysis: Empirical Validation of the Control Knobs}
As aforementioned in Section~\ref{method:APReasoner}, \ap{} is equipped with two explicit control knobs: $\alpha$ for modulating diversity, and $\beta$ for modulating query identity.

For $\alpha$, as illustrated in Table~\ref{tab:topl5_informative_alpha} and Appendix~\ref{si:alpha-result}, experiment results indicate that long range contact prediction average performance improves as $\alpha$ grows, consistent with the initial design idea that larger $\alpha$ promotes more diverse AP-selected subsets.

\begin{figure}[t]
  \centering
  \noindent\makebox[\linewidth][c]{%
    \includegraphics[width=0.9\linewidth]{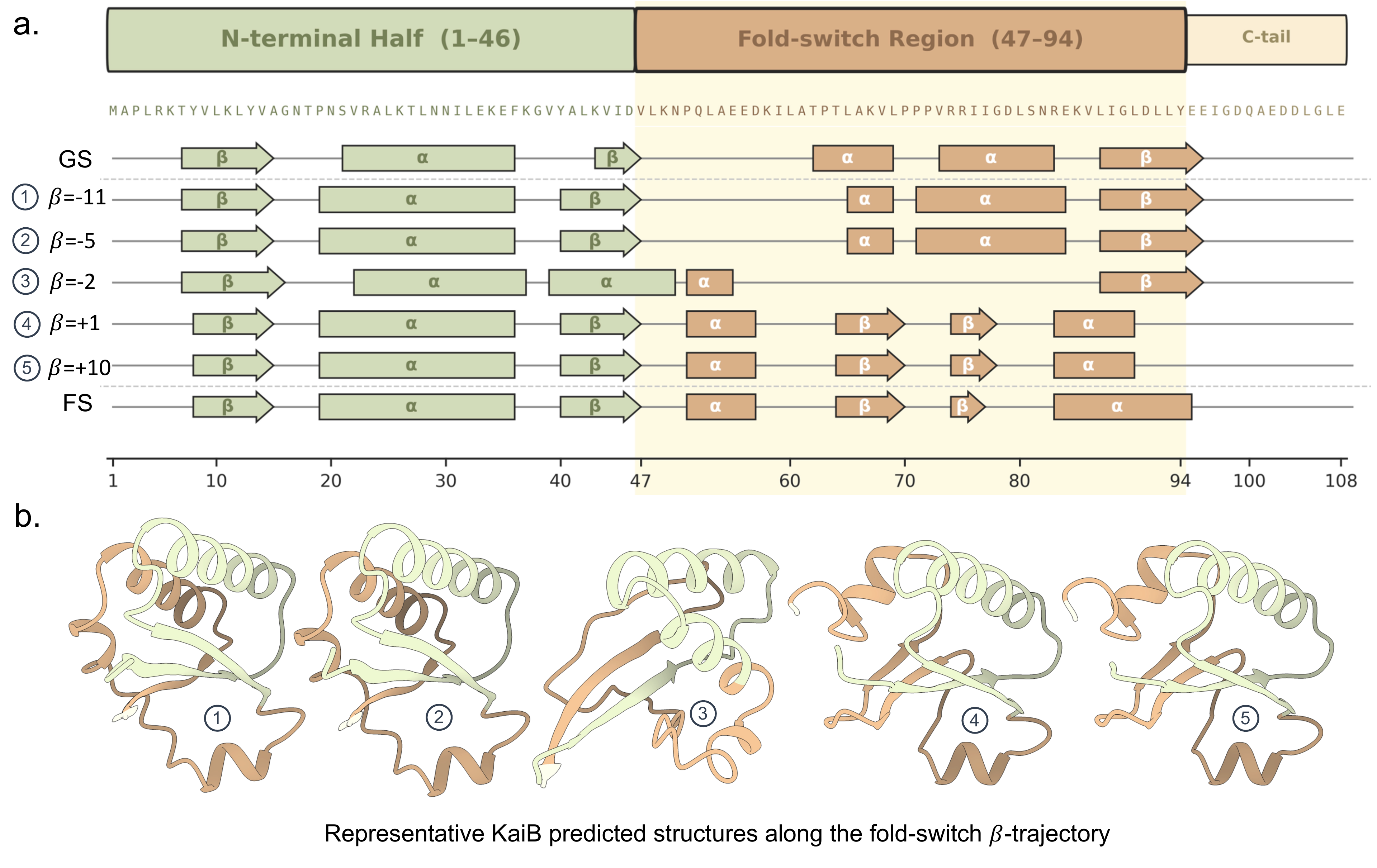}%
  }
\caption{Structural transition of KaiB controlled by the $\beta$ knob. (a) Secondary structure evolution of the fold-switch region (residues 47--94), showing the transition from Ground State (GS) topology at $\beta = -11$ to the Fold-switch (FS) state at $\beta = +10$. (b) Representative predicted snapshots (1--5) along the transition trajectory, capturing the continuous reconfiguration and biologically relevant intermediates between the two functional basins.}
\label{fig:pmcp_alpha}
  \vspace{-5pt}
\end{figure}

\begin{figure}[t]
  \centering
  \noindent\makebox[\linewidth][c]{%
    \includegraphics[width=0.9\linewidth]{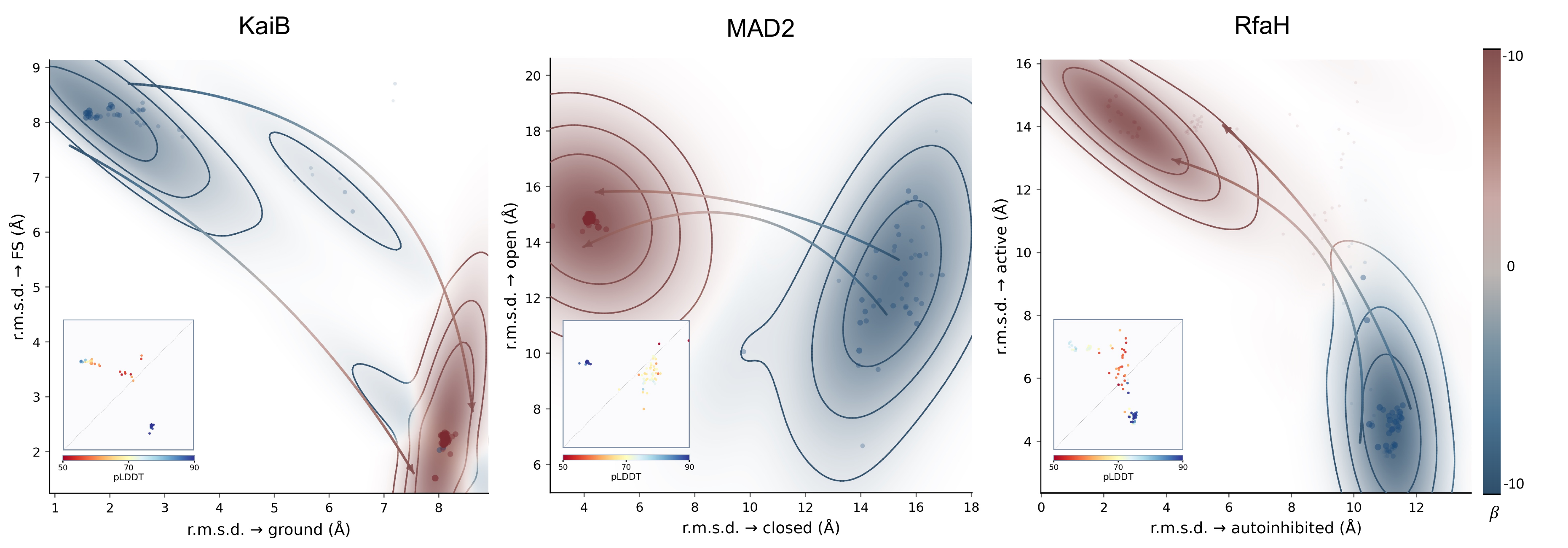}%
  }
  \caption{r.m.s.d. of protein structure predictions by \ap{}. With the control knob $\beta$ increasing, the predictions move from one high confidence region to another one, indicating that \ap{} can help find different conformations by changing control knobs $\beta$. }
  \label{fig:knob_beta}
  \vspace{-15pt}
\end{figure}
For the empirical validation of the control knobs, as illustrated in Figure \ref{fig:pmcp_alpha} and Figure \ref{fig:knob_beta}, the $\beta$ parameter acts as a directional steering mechanism that modulates the model's focus within the complex conformational landscape. By varying $\beta$ across a range from $-10$ to $+10$, the predicted ensemble effectively shifts between distinct functional basins. These include the Ground and FS states in KaiB, the Closed and Open states in MAD2, and the Autoinhibited and Active states in RfaH. This demonstrates a clear and controllable transition trajectory that allows for the predictable recovery of specific functional states that are often lost or underrepresented in stochastic heuristic sampling. This controllable behavior is further validated by the secondary structure reconfiguration observed in KaiB. As shown in Figure \ref{fig:pmcp_alpha}, at low $\beta$ values such as $\beta = -11$, the fold-switch region covering residues 47 to 94 accurately maintains the Ground State topology. As $\beta$ increases towards $+10$, the region undergoes a systematic and dramatic reconfiguration into the $\beta$-sheet and $\alpha$-helix motifs characteristic of the FS state. By enabling the targeted recovery of these specific states and their structural intermediates visible in the representative snapshots 1 to 5, the $\beta$ knob proves that \ap{} can successfully deconvolve overlapping evolutionary signals latent within the MSA. Importantly, this controllable transition is not unique to KaiB. Consistent results for RfaH and MAD2 provided in Figure \ref{fig:RafH_SI} and Figure \ref{fig:MAD_SI} further confirm the generalizability of the $\beta$ knob in steering transitions across diverse fold-switching architectures. This provides a resource-efficient alternative to brute-force methods, allowing for high-fidelity modeling of the protein's full functional ensemble without the prohibitive computational overhead of random subsampling.

\section{Conclusion}
In this paper, we recast MSA subsampling from a heuristic procedure into an explicit optimization problem, and propose \ap{}, an Affinity-Propagation-based factor-graph approach that coordinates query identity and sequence diversity through factor design reasoning. Two interpretable control knobs allow these evolutionary signals to be modulated. Experiments on long-range contact prediction and conformational ensemble prediction show that \ap{} outperforms naive and heuristic baselines, and enables controllable recovery of alternative functional states in KaiB, MAD2, and RfaH. These results demonstrate the advantage of treating MSA subsampling as a controllable optimization problem solved through factor-graph reasoning. Despite the effectiveness of \ap{}, several limitations remain in terms of computational efficiency and large-scale validation. First, \ap{} introduces additional computational cost from AP message passing and the fixed-cardinality wrapper, and is approximately 2--3$\times$ slower than lightweight heuristic samplers such as divMax, divMin, and HHfilter under the same data-scale settings. 
Second, due to computational constraints, we train on a 260K-scale dataset rather than the full 26M-scale corpus, so the effectiveness and scalability of \ap{} in substantially larger pretraining regimes remain to be further explored.

\medskip


\newpage

\section{Broader Impact} The implications of this work are significant for both fundamental biology and therapeutic development. By enabling the high-fidelity recovery of specific functional basins, such as the active and inhibited states of proteins like RfaH and MAD2, \ap{} provides a resource-efficient tool for investigating protein dynamics and allosteric mechanisms. This capability is crucial for the design of "smart" therapeutics that target specific conformational states, as well as for the advancement of protein engineering, where understanding structural plasticity is essential for developing novel enzymes and biosensors. Furthermore, the steerability of our model reduces the computational overhead associated with brute-force sampling, offering a scalable and predictable pathway for large-scale structural genomics and the characterization of the dark proteome.

\bibliographystyle{plainnat}
\bibliography{references}

@article{rives2021biological,
  title={Biological structure and function emerge from scaling unsupervised learning to 250 million protein sequences},
  author={Rives, Alexander and Meier, Joshua and Sercu, Tom and Goyal, Siddharth and Lin, Zeming and Liu, Jason and Guo, Demi and Ott, Myle and Zitnick, C Lawrence and Ma, Jerry and others},
  journal={Proceedings of the national academy of sciences},
  volume={118},
  number={15},
  pages={e2016239118},
  year={2021},
  publisher={National Academy of Sciences}
}

@article{heinzinger2019modeling,
  title={Modeling aspects of the language of life through transfer-learning protein sequences},
  author={Heinzinger, Michael and Elnaggar, Ahmed and Wang, Yu and Dallago, Christian and Nechaev, Dmitrii and Matthes, Florian and Rost, Burkhard},
  journal={BMC bioinformatics},
  volume={20},
  number={1},
  pages={723},
  year={2019},
  publisher={Springer}
}

@article{vig2020bertology,
  title={Bertology meets biology: Interpreting attention in protein language models},
  author={Vig, Jesse and Madani, Ali and Varshney, Lav R and Xiong, Caiming and Socher, Richard and Rajani, Nazneen Fatema},
  journal={arXiv preprint arXiv:2006.15222},
  year={2020}
}

@inproceedings{notin2022tranception,
  title={Tranception: protein fitness prediction with autoregressive transformers and inference-time retrieval},
  author={Notin, Pascal and Dias, Mafalda and Frazer, Jonathan and Marchena-Hurtado, Javier and Gomez, Aidan N and Marks, Debora and Gal, Yarin},
  booktitle={International Conference on Machine Learning},
  pages={16990--17017},
  year={2022},
  organization={PMLR}
}

@article{lin2023evolutionary,
  title={Evolutionary-scale prediction of atomic-level protein structure with a language model},
  author={Lin, Zeming and Akin, Halil and Rao, Roshan and Hie, Brian and Zhu, Zhongkai and Lu, Wenting and Smetanin, Nikita and Verkuil, Robert and Kabeli, Ori and Shmueli, Yaniv and others},
  journal={Science},
  volume={379},
  number={6637},
  pages={1123--1130},
  year={2023},
  publisher={American Association for the Advancement of Science}
}

@article{del2022sampling,
  title={Sampling alternative conformational states of transporters and receptors with AlphaFold2},
  author={Del Alamo, Diego and Sala, Davide and Mchaourab, Hassane S and Meiler, Jens},
  journal={elife},
  volume={11},
  pages={e75751},
  year={2022},
  publisher={eLife Sciences Publications, Ltd}
}

@article{wayment2024predicting,
  title={Predicting multiple conformations via sequence clustering and AlphaFold2},
  author={Wayment-Steele, Hannah K and Ojoawo, Adedolapo and Otten, Renee and Apitz, Julia M and Pitsawong, Warintra and H{\"o}mberger, Marc and Ovchinnikov, Sergey and Colwell, Lucy and Kern, Dorothee},
  journal={Nature},
  volume={625},
  number={7996},
  pages={832--839},
  year={2024},
  publisher={Nature Publishing Group UK London}
}

@inproceedings{rao2021msa,
  title={MSA transformer},
  author={Rao, Roshan M and Liu, Jason and Verkuil, Robert and Meier, Joshua and Canny, John and Abbeel, Pieter and Sercu, Tom and Rives, Alexander},
  booktitle={International conference on machine learning},
  pages={8844--8856},
  year={2021},
  organization={PMLR}
}

@article{abramson2024accurate,
  title={Accurate structure prediction of biomolecular interactions with AlphaFold 3},
  author={Abramson, Josh and Adler, Jonas and Dunger, Jack and Evans, Richard and Green, Tim and Pritzel, Alexander and Ronneberger, Olaf and Willmore, Lindsay and Ballard, Andrew J and Bambrick, Joshua and others},
  journal={Nature},
  volume={630},
  number={8016},
  pages={493--500},
  year={2024},
  publisher={Nature Publishing Group UK London}
}

@article{mirdita2022colabfold,
  title={ColabFold: making protein folding accessible to all},
  author={Mirdita, Milot and Sch{\"u}tze, Konstantin and Moriwaki, Yoshitaka and Heo, Lim and Ovchinnikov, Sergey and Steinegger, Martin},
  journal={Nature methods},
  volume={19},
  number={6},
  pages={679--682},
  year={2022},
  publisher={Nature Publishing Group US New York}
}

@article{steinegger2019hh,
  title={HH-suite3 for fast remote homology detection and deep protein annotation},
  author={Steinegger, Martin and Meier, Markus and Mirdita, Milot and V{\"o}hringer, Harald and Haunsberger, Stephan J and S{\"o}ding, Johannes},
  journal={BMC bioinformatics},
  volume={20},
  number={1},
  pages={473},
  year={2019},
  publisher={Springer}
}

@article{hong2022prot,
  title={A-Prot: protein structure modeling using MSA transformer},
  author={Hong, Yiyu and Lee, Juyong and Ko, Junsu},
  journal={BMC bioinformatics},
  volume={23},
  number={1},
  pages={93},
  year={2022},
  publisher={Springer}
}

@book{pearl2014probabilistic,
  title={Probabilistic reasoning in intelligent systems: networks of plausible inference},
  author={Pearl, Judea},
  year={2014},
  publisher={Elsevier}
}

@article{kschischang2001factor,
  title={Factor graphs and the sum-product algorithm},
  author={Kschischang, Frank R and Frey, Brendan J and Loeliger, H-A},
  journal={IEEE Transactions on information theory},
  volume={47},
  number={2},
  pages={498--519},
  year={2001},
  publisher={IEEE}
}

@article{wainwright2008graphical,
  title={Graphical models, exponential families, and variational inference},
  author={Wainwright, Martin J and Jordan, Michael I},
  journal={Foundations and Trends{\textregistered} in Machine Learning},
  volume={1},
  number={1-2},
  pages={1--305},
  year={2008},
  publisher={Emerald Publishing Limited}
}

@article{krahenbuhl2011efficient,
  title={Efficient inference in fully connected crfs with gaussian edge potentials},
  author={Kr{\"a}henb{\"u}hl, Philipp and Koltun, Vladlen},
  journal={Advances in neural information processing systems},
  volume={24},
  year={2011}
}

@article{huang2015bidirectional,
  title={Bidirectional LSTM-CRF models for sequence tagging},
  author={Huang, Zhiheng and Xu, Wei and Yu, Kai},
  journal={arXiv preprint arXiv:1508.01991},
  year={2015}
}

@article{lafferty2001conditional,
  title={Conditional random fields: Probabilistic models for segmenting and labeling sequence data},
  author={Lafferty, John and McCallum, Andrew and Pereira, Fernando CN},
  year={2001}
}

@inproceedings{NIPS2005_327708dd,
 author = {Frey, Brendan J and Dueck, Delbert},
 booktitle = {Advances in Neural Information Processing Systems},
 editor = {Y. Weiss and B. Sch\"{o}lkopf and J. Platt},
 pages = {},
 publisher = {MIT Press},
 title = {Mixture Modeling by Affinity Propagation},
 url = {https://proceedings.neurips.cc/paper_files/paper/2005/file/327708dd10d68b1361ad3addbaca01f2-Paper.pdf},
 volume = {18},
 year = {2005}
}

@article{frey2007clustering,
author = {Brendan J. Frey  and Delbert Dueck },
title = {Clustering by Passing Messages Between Data Points},
journal = {Science},
volume = {315},
number = {5814},
pages = {972-976},
year = {2007},
doi = {10.1126/science.1136800},
URL = {https://www.science.org/doi/abs/10.1126/science.1136800},
eprint = {https://www.science.org/doi/pdf/10.1126/science.1136800}}

@article{pedregosa2011scikit,
  title={Scikit-learn: Machine learning in Python},
  author={Pedregosa, Fabian and Varoquaux, Ga{\"e}l and Gramfort, Alexandre and Michel, Vincent and Thirion, Bertrand and Grisel, Olivier and Blondel, Mathieu and Prettenhofer, Peter and Weiss, Ron and Dubourg, Vincent and others},
  journal={the Journal of machine Learning research},
  volume={12},
  pages={2825--2830},
  year={2011},
  publisher={JMLR. org}
}

@article{morcos2011direct,
  title={Direct-coupling analysis of residue coevolution captures native contacts across many protein families},
  author={Morcos, Faruck and Pagnani, Andrea and Lunt, Bryan and Bertolino, Arianna and Marks, Debora S and Sander, Chris and Zecchina, Riccardo and Onuchic, Jos{\'e} N and Hwa, Terence and Weigt, Martin},
  journal={Proceedings of the National Academy of Sciences},
  volume={108},
  number={49},
  pages={E1293--E1301},
  year={2011},
  publisher={National Academy of Sciences}
}

@article{tape,
  author       = {Roshan Rao and
                  Nicholas Bhattacharya and
                  Neil Thomas and
                  Yan Duan and
                  Xi Chen and
                  John F. Canny and
                  Pieter Abbeel and
                  Yun S. Song},
  title        = {Evaluating Protein Transfer Learning with {TAPE}},
  journal      = {CoRR},
  volume       = {abs/1906.08230},
  year         = {2019},
  url          = {http://arxiv.org/abs/1906.08230},
  eprinttype   = {arXiv},
  eprint       = {1906.08230},
  bibsource    = {dblp computer science bibliography, https://dblp.org}
}

@article{bert,
  author       = {Jacob Devlin and
                  Ming{-}Wei Chang and
                  Kenton Lee and
                  Kristina Toutanova},
  title        = {{BERT:} Pre-training of Deep Bidirectional Transformers for Language
                  Understanding},
  journal      = {CoRR},
  volume       = {abs/1810.04805},
  year         = {2018},
  url          = {http://arxiv.org/abs/1810.04805},
  eprinttype   = {arXiv},
  eprint       = {1810.04805},
  bibsource    = {dblp computer science bibliography, https://dblp.org}
}

@misc{ahdritz2023openproteinsettrainingdatastructural,
      title={OpenProteinSet: Training data for structural biology at scale}, 
      author={Gustaf Ahdritz and Nazim Bouatta and Sachin Kadyan and Lukas Jarosch and Daniel Berenberg and Ian Fisk and Andrew M. Watkins and Stephen Ra and Richard Bonneau and Mohammed AlQuraishi},
      year={2023},
      eprint={2308.05326},
      archivePrefix={arXiv},
      primaryClass={q-bio.BM},
      url={https://arxiv.org/abs/2308.05326}, 
}

@article{rcsb,
    author = {Hofmann, A. and Wlodawer, A.},
    title = {PCSB—a program collection for structural biology and
  biophysical chemistry },
    journal = {Bioinformatics},
    volume = {18},
    number = {1},
    pages = {209-210},
    year = {2002},
    month = {01},
    issn = {1367-4803},
    doi = {10.1093/bioinformatics/18.1.209},
    url = {https://doi.org/10.1093/bioinformatics/18.1.209},
    eprint = {https://academic.oup.com/bioinformatics/article-pdf/18/1/209/48850395/bioinformatics_18_1_209.pdf},
}

@article{casp15,
  author       = {Kryshtafovych, Andriy and Antczak, Maciej and Szachniuk, Marta and Zok, Tomasz and Kretsch, Rachael C. and Rangan, Ramya and Pham, Phillip and Das, Rhiju and Robin, Xavier and Studer, Gabriel and Durairaj, Janani and Eberhardt, Jerome and Sweeney, Aaron and Topf, Maya and Schwede, Torsten and Fidelis, Krzysztof and Moult, John},
  title        = {New prediction categories in CASP15},
  journal      = {Proteins: Structure, Function and Bioinformatics},
  volume       = {91},
  number       = {12},
  pages        = {1550--1557},
  year         = {2023},
  doi          = {10.1002/prot.26515},
}

@article{jumper2021highly,
  title     = {Highly accurate protein structure prediction with AlphaFold},
  author    = {Jumper, John and Evans, Richard and Pritzel, Alexander and Green, Tim and Figurnov, Michael and Ronneberger, Olaf and Tunyasuvunakool, Kathryn and Bates, Russ and {\v{Z}}{\'i}dek, Augustin and Potapenko, Anna and Bridgland, Alex and Meyer, Clemens and Kohl, Simon A. A. and Ballard, Andrew J. and Cowie, Andrew and Romera-Paredes, Bernardino and Nikolov, Stanislav and Jain, Rishub and Adler, Jonas and Back, Trevor and Petersen, Stig and Reiman, David and Clancy, Ellen and Zielinski, Michal and Steinegger, Martin and Pacholska, Michalina and Berghammer, Tamas and Bodenstein, Sebastian and Silver, David and Vinyals, Oriol and Senior, Andrew W. and Kavukcuoglu, Koray and Kohli, Pushmeet and Hassabis, Demis},
  journal   = {Nature},
  volume    = {596},
  number    = {7873},
  pages     = {583--589},
  year      = {2021},
  doi       = {10.1038/s41586-021-03819-2},
  publisher = {Nature Publishing Group}
}

@inproceedings{qu2009collaborative,
    title = "Collaborative Summarization: When Collaborative Filtering Meets Document Summarization",
    author = "Qu, Yang  and
      Chen, Qunxiu",
    editor = "Kwong, Olivia",
    booktitle = "Proceedings of the 23rd Pacific Asia Conference on Language, Information and Computation, Volume 2",
    month = dec,
    year = "2009",
    address = "Hong Kong",
    publisher = "City University of Hong Kong",
    url = "https://aclanthology.org/Y09-2005/",
    pages = "474--483"
}

@article{leone2007clustering,
   title={Clustering by soft-constraint affinity propagation: applications to gene-expression data},
   volume={23},
   ISSN={1367-4803},
   url={http://dx.doi.org/10.1093/bioinformatics/btm414},
   DOI={10.1093/bioinformatics/btm414},
   number={20},
   journal={Bioinformatics},
   publisher={Oxford University Press (OUP)},
   author={Leone, Michele and Sumedha and Weigt, Martin},
   year={2007},
   month=Sept, pages={2708–2715} }

@inproceedings{DBLP:conf/iclr/RaoMSOR21,
  author       = {Roshan Rao and
                  Joshua Meier and
                  Tom Sercu and
                  Sergey Ovchinnikov and
                  Alexander Rives},
  title        = {Transformer protein language models are unsupervised structure learners},
  booktitle    = {9th International Conference on Learning Representations, {ICLR} 2021,
                  Virtual Event, Austria, May 3-7, 2021},
  publisher    = {OpenReview.net},
  year         = {2021},
  url          = {https://openreview.net/forum?id=fylclEqgvgd},
  bibsource    = {dblp computer science bibliography, https://dblp.org}
}

@article{steinegger2017mmseqs2,
  author  = {Steinegger, Martin and S{\"o}ding, Johannes},
  title   = {{MMseqs2} enables sensitive protein sequence searching for the analysis of massive data sets},
  journal = {Nature Biotechnology},
  year    = {2017},
  volume  = {35},
  number  = {11},
  pages   = {1026--1028},
  doi     = {10.1038/nbt.3988},
  url     = {https://doi.org/10.1038/nbt.3988}
}

@article{chothia1986relation,
  author  = {Chothia, Cyrus and Lesk, Arthur M.},
  title   = {The relation between the divergence of sequence and structure in proteins},
  journal = {The EMBO Journal},
  year    = {1986},
  volume  = {5},
  number  = {4},
  pages   = {823--826},
  doi     = {10.1002/j.1460-2075.1986.tb04288.x},
  pmid    = {3709526},
  pmcid   = {PMC1166865}
}

@article{anfinsen1973principles,
  title={Principles that govern the folding of protein chains},
  author={Anfinsen, Christian B},
  journal={Science},
  volume={181},
  number={4096},
  pages={223--230},
  year={1973},
  publisher={American Association for the Advancement of Science}
}

@article{dill2012protein,
  title={The protein-folding problem, 50 years on},
  author={Dill, Ken A and MacCallum, Justin L},
  journal={science},
  volume={338},
  number={6110},
  pages={1042--1046},
  year={2012},
  publisher={American Association for the Advancement of Science}
}

@article{senior2020improved,
  title={Improved protein structure prediction using potentials from deep learning},
  author={Senior, Andrew W and Evans, Richard and Jumper, John and Kirkpatrick, James and Sifre, Laurent and Green, Tim and Qin, Chongli and {\v{Z}}{\'\i}dek, Augustin and Nelson, Alexander WR and Bridgland, Alex and others},
  journal={Nature},
  volume={577},
  number={7792},
  pages={706--710},
  year={2020},
  publisher={Nature Publishing Group UK London}
}

@article{baek2021accurate,
  title={Accurate prediction of protein structures and interactions using a three-track neural network},
  author={Baek, Minkyung and DiMaio, Frank and Anishchenko, Ivan and Dauparas, Justas and Ovchinnikov, Sergey and Lee, Gyu Rie and Wang, Jue and Cong, Qian and Kinch, Lisa N and Schaeffer, R Dustin and others},
  journal={Science},
  volume={373},
  number={6557},
  pages={871--876},
  year={2021},
  publisher={American Association for the Advancement of Science}
}

@article{evans2021protein,
  title={Protein complex prediction with AlphaFold-Multimer},
  author={Evans, Richard and O’neill, Michael and Pritzel, Alexander and Antropova, Natasha and Senior, Andrew and Green, Tim and {\v{Z}}{\'\i}dek, Augustin and Bates, Russ and Blackwell, Sam and Yim, Jason and others},
  journal={biorxiv},
  pages={2021--10},
  year={2021},
  publisher={Cold Spring Harbor Laboratory}
}

@article{yang2020improved,
  title={Improved protein structure prediction using predicted interresidue orientations},
  author={Yang, Jianyi and Anishchenko, Ivan and Park, Hahnbeom and Peng, Zhenling and Ovchinnikov, Sergey and Baker, David},
  journal={Proceedings of the National Academy of Sciences},
  volume={117},
  number={3},
  pages={1496--1503},
  year={2020},
  publisher={National Academy of Sciences}
}

@article{wang2017accurate,
  title={Accurate de novo prediction of protein contact map by ultra-deep learning model},
  author={Wang, Sheng and Sun, Siqi and Li, Zhen and Zhang, Renyu and Xu, Jinbo},
  journal={PLoS computational biology},
  volume={13},
  number={1},
  pages={e1005324},
  year={2017},
  publisher={Public Library of Science San Francisco, CA USA}
}

@article{marks2011protein,
  title={Protein 3D structure computed from evolutionary sequence variation},
  author={Marks, Debora S and Colwell, Lucy J and Sheridan, Robert and Hopf, Thomas A and Pagnani, Andrea and Zecchina, Riccardo and Sander, Chris},
  journal={PloS one},
  volume={6},
  number={12},
  pages={e28766},
  year={2011},
  publisher={Public Library of Science San Francisco, USA}
}

@article{hopf2017mutation,
  title={Mutation effects predicted from sequence co-variation},
  author={Hopf, Thomas A and Ingraham, John B and Poelwijk, Frank J and Sch{\"a}rfe, Charlotta PI and Springer, Michael and Sander, Chris and Marks, Debora S},
  journal={Nature biotechnology},
  volume={35},
  number={2},
  pages={128--135},
  year={2017},
  publisher={Nature Publishing Group US New York}
}

@article{frazer2021disease,
  title={Disease variant prediction with deep generative models of evolutionary data},
  author={Frazer, Jonathan and Notin, Pascal and Dias, Mafalda and Gomez, Aidan and Min, Joseph K and Brock, Kelly and Gal, Yarin and Marks, Debora S},
  journal={Nature},
  volume={599},
  number={7883},
  pages={91--95},
  year={2021},
  publisher={Nature Publishing Group UK London}
}

@article{SPEACHAF,
    doi = {10.1371/journal.pcbi.1010483},
    author = {Stein, Richard A. AND Mchaourab, Hassane S.},
    journal = {PLOS Computational Biology},
    publisher = {Public Library of Science},
    title = {SPEACH\_AF: Sampling protein ensembles and conformational heterogeneity with Alphafold2},
    year = {2022},
    month = {08},
    volume = {18},
    url = {https://doi.org/10.1371/journal.pcbi.1010483},
    pages = {1-16},
    number = {8},

}

@article{elnaggar2021prottrans,
  title={ProtTrans: toward understanding the language of life through self-supervised learning},
  author={Elnaggar, Ahmed and Heinzinger, Michael and Dallago, Christian and Rehawi, Ghalia and Wang, Yu and Jones, Llion and Gibbs, Tom and Feher, Tamas and Angerer, Christoph and Steinegger, Martin and others},
  journal={IEEE transactions on pattern analysis and machine intelligence},
  volume={44},
  number={10},
  pages={7112--7127},
  year={2021},
  publisher={IEEE}
}

@article{madani2023large,
  title={Large language models generate functional protein sequences across diverse families},
  author={Madani, Ali and Krause, Ben and Greene, Eric R and Subramanian, Subu and Mohr, Benjamin P and Holton, James M and Olmos Jr, Jose Luis and Xiong, Caiming and Sun, Zachary Z and Socher, Richard and others},
  journal={Nature biotechnology},
  volume={41},
  number={8},
  pages={1099--1106},
  year={2023},
  publisher={Nature Publishing Group US New York}
}

@article{MonteiroDaSilva2024SubsampledAlphaFold2,
  author  = {Monteiro da Silva, Gabriel and Cui, Jennifer Y. and Dalgarno, David C. and Lisi, George P. and Rubenstein, Brenda M.},
  title   = {High-throughput prediction of protein conformational distributions with subsampled {AlphaFold2}},
  journal = {Nature Communications},
  year    = {2024},
  volume  = {15},
  number  = {1},
  pages   = {2464},
  doi     = {10.1038/s41467-024-46715-9},
  url     = {https://doi.org/10.1038/s41467-024-46715-9},
  issn    = {2041-1723}
}

@article{Lee2025CFRandom,
  author  = {Lee, Myeongsang and Schafer, Joseph W. and Prabakaran, Jeshuwin and Chakravarty, Devlina and Clore, Madeleine F. and Porter, Lauren L.},
  title   = {Large-scale predictions of alternative protein conformations by {AlphaFold2}-based sequence association},
  journal = {Nature Communications},
  year    = {2025},
  volume  = {16},
  number  = {1},
  pages   = {5622},
  doi     = {10.1038/s41467-025-60759-5},
  url     = {https://doi.org/10.1038/s41467-025-60759-5},
  issn    = {2041-1723}
}

@article{stein2022speach_af,
  title={SPEACH\_AF: Sampling protein ensembles and conformational heterogeneity with Alphafold2},
  author={Stein, Richard A and Mchaourab, Hassane S},
  journal={PLoS computational biology},
  volume={18},
  number={8},
  pages={e1010483},
  year={2022},
  publisher={Public Library of Science San Francisco, CA USA}
}

@article{petti2023end,
  title={End-to-end learning of multiple sequence alignments with differentiable Smith--Waterman},
  author={Petti, Samantha and Bhattacharya, Nicholas and Rao, Roshan and Dauparas, Justas and Thomas, Neil and Zhou, Juannan and Rush, Alexander M and Koo, Peter and Ovchinnikov, Sergey},
  journal={Bioinformatics},
  volume={39},
  number={1},
  pages={btac724},
  year={2023},
  publisher={Oxford University Press}
}


\appendix
\newpage

\startcontents[appendix]

\section*{Contents of the Appendix}
\printcontents[appendix]{}{1}{}
\newpage

\section{\ap{} via Factor Graph}
\subsection{From Subsampling to Reasoning}
The MSA subsampling for MSA-based Protein Language Model Training is an important and realistic procedure from the perspective of computational limit. For \ap{}s, the final output remains a subsampled MSA subset, but the central idea is no longer to treat subsampling as a one-shot filtering problem. Instead, we view it as an MSA reasoning problem, in which the retained rows should be inferred jointly according to how well they represent the full alignment, how much redundancy they remove, etc. The \ap{} redefines subsampling by shifting the paradigm from passive, one-shot sequence filtering to an active, joint inference process that globally optimizes for evolutionary representativeness and information density.

\subsection{Factor Graph and Reasoning}

A factor graph is a bipartite graphical model that represents a global objective or joint distribution by decomposing it into local factors over subsets of variables. Formally, for latent variables \(z_{1:n}\), a factor graph writes
\[
p(z_{1:n}) \propto \prod_{a} f_a(z_a),
\]
where each factor \(f_a\) scores or constrains only a small subset \(z_a\) of the variables. This representation differs from graph neural networks (GNNs) in an important way: a factor graph is not primarily a learned feature-propagation architecture, but an explicit structured model whose nodes and edges have predefined semantic roles. Variable nodes represent latent choices, factor nodes encode compatibility or consistency constraints, and message passing is derived from the underlying objective itself rather than learned as a generic parametric operator. In this sense, the intermediate inference process on a factor graph can be regarded as reasoning, because it is an iterative procedure that combines local evidence with global structural constraints to infer coherent latent assignments. 

\section{Additional details of \ap{}}

\subsection{Relation to Vanilla Affinity Propagation}
Affinity Propagation is well suited for MSA subsampling because its goal of selecting exemplars directly matches the need to choose representative sequences from an MSA, while its message-passing updates coordinate these choices over all candidate rows. Extensive applications across diverse domains, including face-image clustering, document summarization and gene-expression analysis, have established AP as a versatile paradigm~\citep{frey2007clustering, qu2009collaborative, leone2007clustering}. 

\ap{} is an adaptation of AP to fixed-budget MSA subsampling. The underlying AP factor graph and the max-product message-passing schedule are unchanged from the standard formulation. Our specific modeling choices and algorithmic adaptations to the standard AP framework are detailed as follows:
\begin{enumerate}
\item \emph{MSA-tailored factor formulation}: We construct the factors specifically for the MSA reasoning scenario as elaborated in Section~\ref{method:APReasoner}. 
\item \emph{Hard query anchoring}: We force the primary query sequence to act as its own exemplar by assigning it an overwhelmingly large preference ($S_{00} = p_q$).
\item \emph{Exact-budget search wrapper}: Because protein language model training generally favors a fixed sequence cardinality to maximize the utilization of the limited token budget, we wrap the AP reasoning in an outer-loop bisection. This mechanism iteratively adjusts the global baseline preference $p$ to guide AP's exemplar count towards the target.
\end{enumerate}

\subsection{Why \(\alpha\) is an exponent rather than a linear coefficient}

We use \(\alpha\) as an exponent because, according to~\citet{frey2007clustering}, a simple multiplicative coefficient on distance would mostly induce a uniform rescaling of all off-diagonal similarities, which can be largely absorbed by the AP preference level and thus provides little genuine control over the selected exemplar geometry. By contrast, an exponent changes the curvature of the distance-to-similarity map: \(\alpha>1\) suppresses short-range differences more strongly than long-range ones, while \(0<\alpha<1\) does the opposite, thereby directly controlling local redundancy collapse and the global spread of exemplars.

\subsection{Exemplar-consistency Factor}
\label{si:fk}
Exemplar-consistency factor $f_k$ couples the assignment decisions across rows and rules out inconsistent configurations in which a row represents others without being an exemplar itself. We use the practical exemplar-consistency factor corresponding to the responsibility/availability updates used in \texttt{sklearn}. For each candidate exemplar $k$, the factor $f_k$ enforces the standard AP implication constraint that any row selected as an exemplar by others must also select itself. Formally, $f_k$ may be written in piecewise form as
\begin{equation}
f_k(s_{1:M})
=
\begin{cases}
0, & \text{if } s_k \neq k \text{ and } \exists\, i \neq k \text{ with } s_i = k,\\
1, & \text{otherwise}.
\end{cases}
\end{equation}
Thus, $f_k$ rules out inconsistent configurations in which row $k$ is chosen as an exemplar by other rows without declaring itself an exemplar. Importantly, this definition still allows singleton exemplars and it only excludes assignments that violate exemplar consistency.

\subsection{Log-domain Max-product Reasoning.}
\label{si:logreasoning}
To perform global optimization over the joint distribution, our implementation follows the log-domain max-product form of affinity propagation popularized by~\citet{frey2007clustering} and used by scikit-learn~\citep{pedregosa2011scikit}. The network iteratively updates two complementary types of messages to simulate a global negotiation. The responsibility message,
\begin{equation}
r(i,k) \leftarrow S_{ik} - \max_{k' \neq k} \left[a(i,k') + S_{ik'}\right],
\end{equation}
quantifies the accumulated evidence for row $i$ to choose candidate $k$ as its exemplar, inherently discounting the strongest competing candidate $k'$. Conversely, the availability messages,
\begin{align}
a(k,k) &\leftarrow \sum_{i' \neq k} \max\!\left(0, r(i',k)\right), \\
a(i,k) &\leftarrow \min\!\Bigg( 0,\; r(k,k) + \sum_{i' \notin \{i,k\}} \max\!\left(0, r(i',k)\right) \Bigg), \qquad i \neq k.
\end{align}
broadcast the capacity of row $k$ to represent $i$, which is dynamically driven by the positive support $k$ gathers from other sequences. Further technical details regarding the max-product reasoning process are provided in the Appendix~\ref{si:logmax}.

The process terminates when the inferred exemplar set remains stable for a fixed number of iterations. Ultimately, the final subset assignments are decoded by summing the converged messages:
\begin{equation}
\hat{s}_i = \arg\max_k \left[a(i,k) + r(i,k)\right],
\end{equation}
where the final selected exemplars are precisely the indices satisfying $\hat{s}_k = k$.

\subsection{From Sum-product AP to the Practical Log-domain Max-product Form}
\label{si:logmax}
Because the original probability-domain sum-product updates involve repeated products, ratios, and summations that are numerically inappropriate in finite precision, we adopt the standard practical log-domain max-product (equivalently, max-sum) formulation, consistent with the responsibility/availability updates used in \texttt{sklearn}. Hereby we briefly introduce how the standard probability-domain AP updates approximately reduce to the log-domain max-product form. In the original sum-product view, the responsibility and availability messages satisfy
\begin{align}
r_{ik}
&\leftarrow
\frac{L_{ik}}{\sum_{j \neq k} a_{ij} L_{ij}},
\\
a_{kk}
&\leftarrow
\prod_{j \neq k} (1+r_{jk}) - 1,
\\
a_{ki}
&\leftarrow
\frac{1}{(1/r_{kk})P_i + 1 - P_i},
\qquad
P_i = \prod_{j \notin \{k,i\}} (1+r_{jk})^{-1},
\end{align}
where $L_{ik} = \exp(S_{ik})$. Taking logs, defining $\widetilde{r}(i,k) = \log r_{ik}$ and $\widetilde{a}(i,k) = \log a_{ik}$, and applying the standard approximations
\begin{equation}
\log \sum_j \exp(z_j) \approx \max_j z_j,
\qquad
\log(1+\exp(z)) \approx \max(0,z),
\end{equation}
eventually lead to
\begin{align}
\widetilde{r}(i,k)
&=
S_{ik}
-
\max_{k' \neq k}
\left[\widetilde{a}(i,k') + S_{ik'}\right],
\\
\widetilde{a}(k,k)
&=
\sum_{i' \neq k} \max\!\left(0,\widetilde{r}(i',k)\right),
\\
\widetilde{a}(i,k)
&=
\min\!\Bigg(
0,\;
\widetilde{r}(k,k)
+
\sum_{i' \notin \{i,k\}}
\max\!\left(0,\widetilde{r}(i',k)\right)
\Bigg),
\qquad i \neq k.
\end{align}
Dropping the tilde recovers the usual log-domain AP notation used in the implementation. The $\min(0,\cdot)$ form is not an ad hoc clipping trick, but arises naturally under the same log-domain max-product approximation applied to the availability messages. Operationally, it means that off-diagonal availabilities can only discourage unstable exemplars; they cannot provide positive support beyond zero.

\subsection{Fixed-cardinality Wrapper}
\label{si:wrapper}
AP controls the number of exemplars implicitly through the self-preference $p$, whereas the protein language modeling training scenario favors an exact token budget and consequently an exact row budget $N_B(L)$. As observed in the original AP work~\citep{NIPS2005_327708dd}, the self-preference level $p$ is empirically monotone with respect to the number of inferred exemplars: larger $p$ typically yields more exemplars. We therefore place AP inside a fixed-cardinality wrapper: we perform bisection over the global preference offset $p$ until the number of inferred exemplars is closest to $N_B(L)$, while keeping the query anchor and any query-dependent bias fixed. If the resulting exemplar count is still above $N$, we retain the $N$ highest-confidence exemplars according to the diagonal belief $a(k,k)+r(k,k)$; if it is below $N$, we pad with the highest-confidence non-exemplars. This yields an exact-$N$ subset without modifying the AP factor graph itself.

A fixed-cardinality wrapper is important for two reasons. First, MSA-based PLMs such as MSA Transformer operate under an explicit token budget. Second, the baseline samplers used throughout this work, including random and divMax, also produce fixed-budget subsets. Matching this interface is therefore necessary for fair comparison. More broadly, controlling the final subset size isolates the effect of subset quality from MSA depth.

Vanilla AP exposes the preference $p$ rather than the target number of exemplars. To obtain an exact row budget $N$, we run AP inside an outer bisection loop over a shared preference offset $p$. We initialize the search at the median off-diagonal similarity, following the standard AP convention used in sklearn, and then expand toward the lower or upper side depending on whether the current exemplar count is above or below $N$. Because the number of AP exemplars is empirically monotone non-decreasing in $p$, bisection can swiftly localize a preference value whose exemplar count is closest to the target budget.

The query-anchor preference $p_q$ and the query-dependent bias term $\beta d_H^{\mathrm{gap}}(k,0)/L$ are held fixed during this search and bisection only adjusts the uniform offset shared by all non-query rows. Importantly, the query-dependent term is treated as a fixed per-row bias, while bisection is performed only over the shared uniform offset \(p\). Thus, all non-query self-preferences are shifted together without changing their relative ordering. This is consistent with the AP paper narrative, in which the diagonal terms encode exemplar-specific priors and therefore need not be identical across rows, so bisection over \(p\) remains a valid and stable mechanism for targeting a desired cardinality even in the presence of fixed row-dependent terms.

\subsection{Implementation Details}
\paragraph{Initialization.}
The assignment variables $s_i$ are not explicitly initialized as hard exemplar choices. Instead, AP is initialized through the similarity matrix $S$, whose diagonal entries encode the self-preferences, together with zero-initialized responsibility and availability messages. Consequently, the first iteration is determined entirely by the relative scores in $S$.

\paragraph{Damping and stopping.}
To prevent oscillations during this iterative reasoning phase, we use damped synchronous updates,
\begin{equation}
m \leftarrow \lambda m_{\mathrm{old}} + (1-\lambda)m_{\mathrm{new}},
\end{equation}
with $\lambda=0.5$. Iteration stops when the inferred exemplar set remains unchanged for 15 consecutive iterations or when a maximum number of iterations is reached.

\paragraph{Complexity.}
The one-off pairwise distance computation costs $\mathcal{O}(M^2L)$. Given the similarity matrix, each AP iteration of responsibility and  availability udpate costs $\mathcal{O}(M^2)$ dense tensor operations. If $T$ denotes the number of inner AP iterations and $B$ the number of bisection steps, the total complexity is therefore $\mathcal{O}(M^2L + BTM^2)$, with $\mathcal{O}(M^2)$ memory. Although this complexity is high, the dense GPU implementation substantially alleviates the practical runtime by parallelizing both the pairwise distance computation and the message-passing updates.

\paragraph{Preprocessing.} 
Because AP is initialized deterministically from the similarity matrix and zero messages, the algorithm is effectively deterministic, and repeated runs produce nearly identical subsampled subsets. Hence we perform AP reasoning offline prior to training. Following MSA Transformer, we cap the input sequence length at 1024 by cropping overlong MSAs before subsampling. Afterwards, the sampler is applied directly to the full MSA during preprocessing and writes out a budget-compliant subset capped at \(\min(N_{\max}, \lfloor B/L \rfloor)\). The downstream PLM is then trained directly on these preprocessed MSAs.

\paragraph{Fixed Hyperparameters} In all training experiments, $\alpha$ and $\beta$ are treated as fixed reasoner hyperparameters rather than learnable model parameters. Their role is not to adapt during optimization, but to prescribe the bias of the reasoning rule before training begins. They serve as explicit control knobs for the diversity and query-identity profile of the resulting MSA subset.


\section{Hump Observation}
\label{si:hump}
\subsection{Metrics}To quantitatively dissect the interplay between sampling strategies and model performance, we define the following metrics to characterize the sampled MSA subsets:\paragraph{Diversity (Coverage).} We employ the average alignment coverage ($\overline{\mathrm{cov}}$) as the primary proxy for the structural diversity and information density of the sampled subset. For a sampled MSA containing $N$ sequences with a query length $L$, it is defined as:\begin{equation}\overline{\mathrm{cov}} = \frac{1}{N \cdot L} \sum_{i=1}^{N} \sum_{j=1}^{L} \mathbb{1}(s_{i,j} \neq \text{'-'})\end{equation}where $\mathbb{1}(\cdot)$ denotes the indicator function. This metric represents the effective sequence breadth provided to the model, which is directly modulated by the diversity control knob $\alpha$ to include varying levels of partially-aligned evolutionary signals.\paragraph{Query Identity (PID).} We measure the evolutionary proximity of the sampled ensemble to the target query using Percentage Identity (PID). Consistent with the decomposition in Eq.~\ref{eq:pid-decomp}, we distinguish between two specific forms:\begin{itemize}\item Column-wise PID ($\mu^{\mathrm{col}}_{\mathrm{PID}}$): The average identity calculated across all $L$ query columns, where gaps are penalized as mismatches. For the $i$-th sequence, it is:\begin{equation}\mathrm{PID}i = \frac{1}{L} \sum{j=1}^{L} \mathbb{1}(s_{i,j} = q_j)\end{equation}where $q_j$ is the $j$-th residue of the query.\item Ungapped PID ($\mu^{\mathrm{ungap}}_{\mathrm{PID}}$): The identity calculated exclusively over the non-gap positions, representing the intrinsic homology of the subjects independent of alignment length:\begin{equation}\mathrm{PID}i^{\mathrm{ungap}} = \frac{\sum{j=1}^{L} \mathbb{1}(s_{i,j} = q_j)}{\sum_{j=1}^{L} \mathbb{1}(s_{i,j} \neq \text{'-'})}\end{equation}\end{itemize}The control knob $\beta$ explicitly modulates $\mu^{\mathrm{col}}_{\mathrm{PID}}$ by filtering sequences based on their identity profiles, effectively steering the model toward specific conformational basins associated with distinct evolutionary clusters. As demonstrated in the "hump" observation, the optimal recovery of rare functional states depends on a critical balance between sequence diversity ($\overline{\mathrm{cov}}$) and query proximity ($\mu^{\mathrm{col}}_{\mathrm{PID}}$), rather than a monotonic increase in sample size.
\subsection{On the Coupling Between Diversity and Query Identity}
\label{si:dq-coupling}

In the OpenProteinSet a3m corpus the query row is gap-free by construction, so for any non-query row $i$ we have $\mathrm{pid}_i = \mathrm{pid}_i^{\mathrm{ungap}} \cdot \mathrm{cov}_i$, since every gap column counts as a mismatch against the gap-free query. Averaging over rows yields
\begin{equation}
\mu^{\mathrm{col}}_{\mathrm{PID}} \;\approx\; \mu^{\mathrm{ungap}}_{\mathrm{PID}} \cdot \overline{\mathrm{cov}}.
\label{eq:pid-decomp}
\end{equation}
On our nine samplers Eq.~\ref{eq:pid-decomp} holds with $R^{2}=0.989$ and a maximum relative error of $5.06\%$. The cross-sampler ranges of the three quantities then pin down the mechanism: $\mu^{\mathrm{col}}_{\mathrm{PID}}$ varies by $3.42\times$, $\mu^{\mathrm{ungap}}_{\mathrm{PID}}$ by only $1.20\times$, and $\overline{\mathrm{cov}}$ by $3.04\times$, with $3.42 \approx 1.20 \times 3.04$. Hence at least $95\%$ of the cross-sampler variance in $\mu^{\mathrm{col}}_{\mathrm{PID}}$ is attributable to coverage rather than residue-level phylogenetic distance, and because MSA-based PLMs treat the gap symbol as an explicit input token, $\mu^{\mathrm{col}}_{\mathrm{PID}}$ is also the input-space query-agreement that the model actually consumes, not a phylogenetic distance.

A direct consequence is that the marginal correlation between diversity and query identity is fully mediated by coverage: $\rho(\mathrm{N_{eff}}/N,\, \mu^{\mathrm{col}}_{\mathrm{PID}})=+0.937$, but $\rho(\mathrm{N_{eff}}/N,\, \mu^{\mathrm{col}}_{\mathrm{PID}}\mid\overline{\mathrm{cov}})=+0.024$ ($n{=}9$) and $-0.020$ ($n{=}8$, excluding \texttt{random}); in parallel, $\rho(\mathrm{N_{eff}}/N,\, \overline{\mathrm{cov}})=+0.958$ already exceeds the marginal correlation against $\mu^{\mathrm{col}}_{\mathrm{PID}}$. Once coverage is held fixed, no independent statistical signal between the two axes remains, and the $(\mathrm{diversity},\, \mu^{\mathrm{col}}_{\mathrm{PID}})$ plane collapses onto a one-dimensional continuum parametrised by coverage. This continuum is precisely the input-space dimension that an MSA-based PLM consumes and on which the hump-shaped law of Section~\ref{exp:lrcp} lives; reporting both metrics is therefore not double-counting evidence but a demonstration that the hump survives reparametrisation and that the $(\alpha,\beta)$ knobs of \textsc{\ap{}} steer one underlying degree of freedom from two complementary directions.

\subsection{Beyond LRCP: Evidence from MLM and Training Dynamics}
\label{si:beyond-lrcp}

\begin{figure*}[h]
    \centering
    \includegraphics[width=\textwidth]{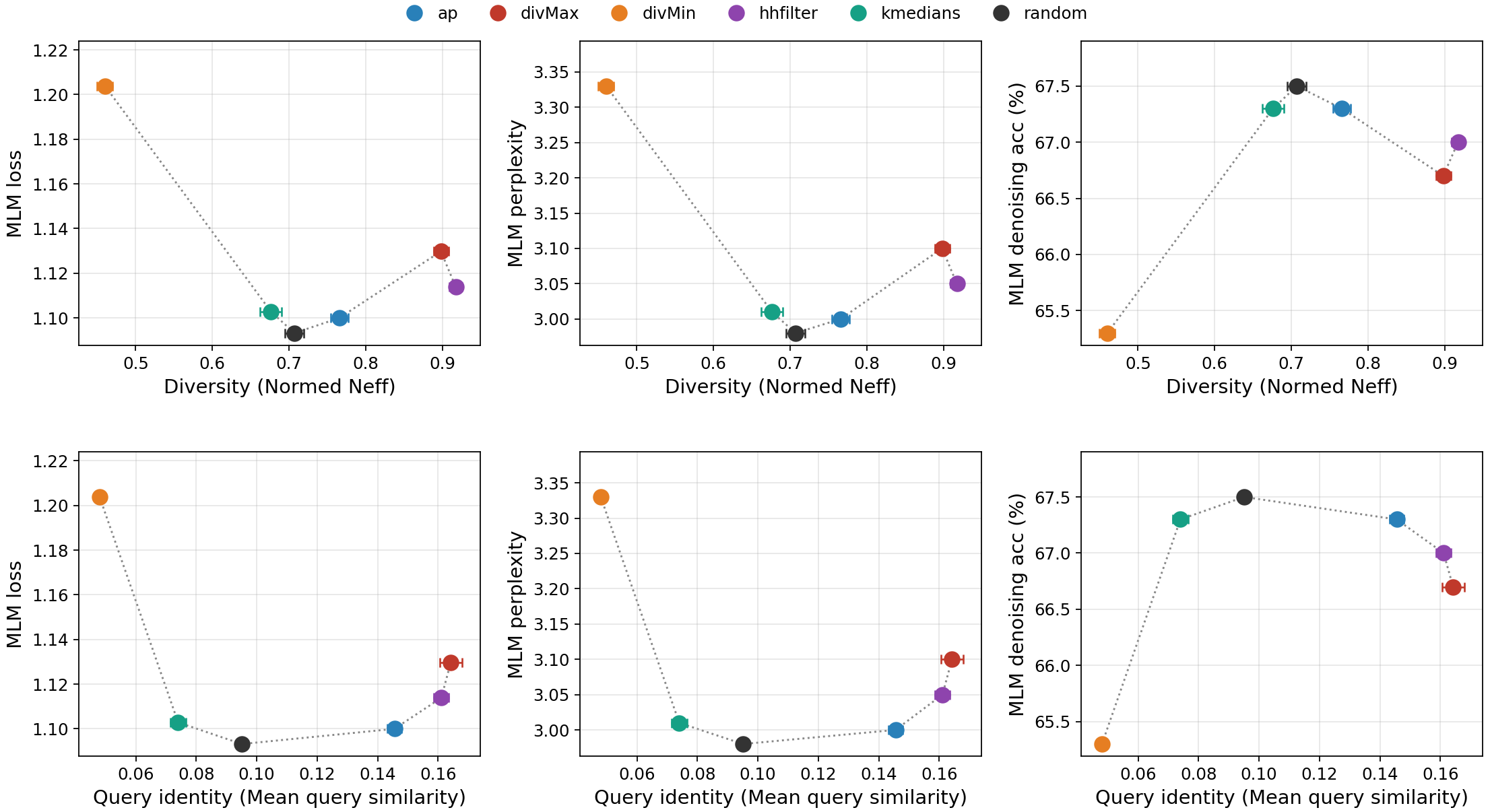}
    \caption{MLM evaluation metrics as a function of sampled-MSA diversity (top row) and mean query similarity (bottom row) across the compared samplers. The best MLM region is concentrated in an intermediate regime rather than at either extreme.}
    \label{fig:mlm-diversity-homology}
\end{figure*}

Although MLM performance is not, by itself, a reliable proxy for downstream structural performance, it still provides a controlled lens for examining how MSA subset statistics affect the learned representation. Figure~\ref{fig:mlm-diversity-homology} shows that MLM behavior varies systematically with both the diversity and the query-identity profile of the sampled MSA. Along the diversity axis, the relationship is clearly non-monotonic. The same qualitative picture appears as well when the x-axis is mean query similarity, indicating that latent property of the sampled MSA can measurably shapes model behavior.

\begin{figure*}[h]
    \centering
    \includegraphics[width=\textwidth]{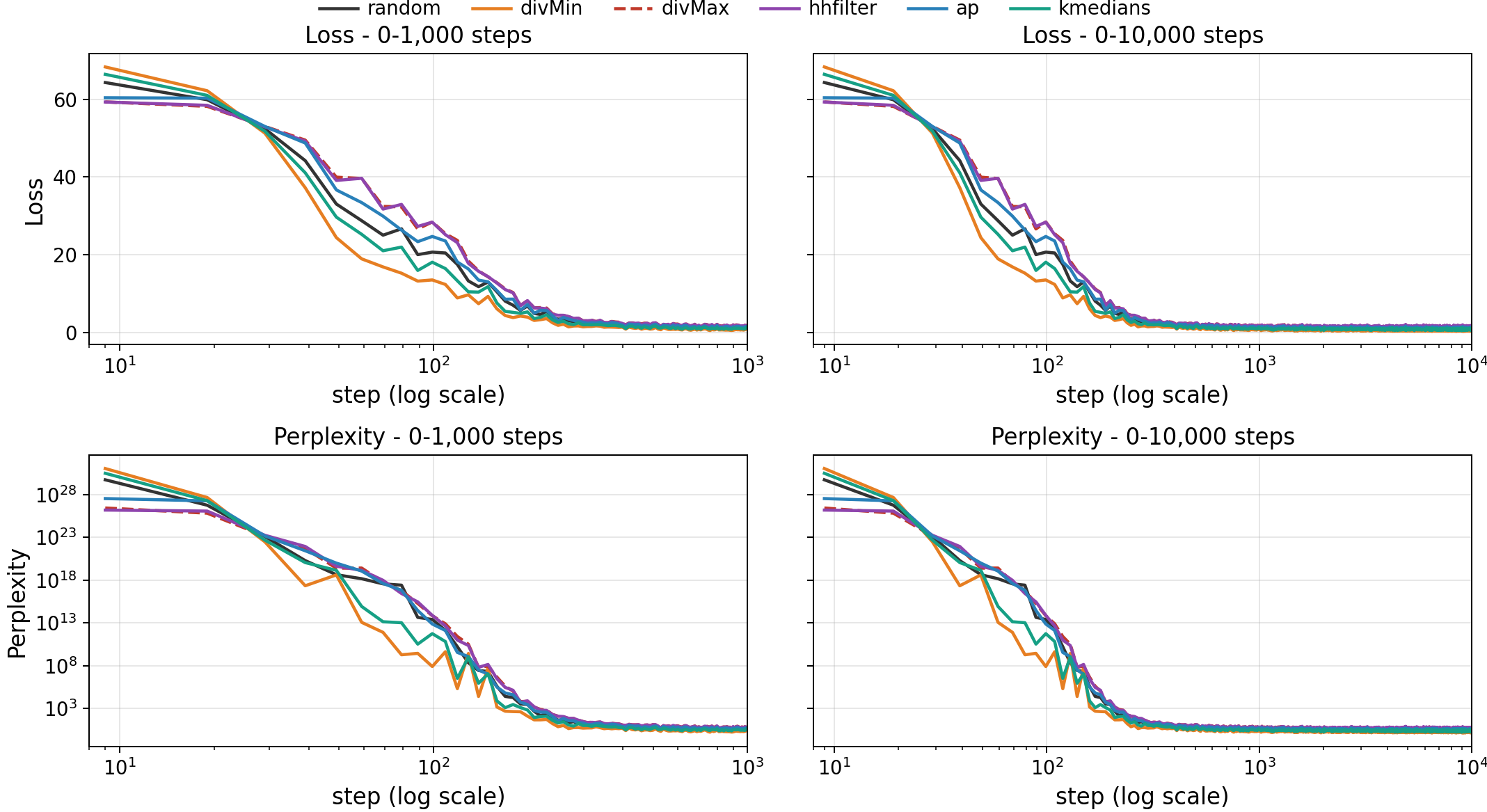}
    \caption{Early training loss and perplexity under different MSA subsampling strategies. Sampler-dependent trajectories diverge early, indicating that subsampling changes the optimization path itself rather than only the final evaluation outcome.}
    \label{fig:early-training-dynamics}
\end{figure*}

Training dynamics provide a second and complementary view. As shown in Figure~\ref{fig:early-training-dynamics}, the loss and perplexity trajectories separate noticeably already in the early stage of training, well before convergence. Different samplers therefore do not merely change the final evaluation numbers; they alter the optimization path itself by changing what evolutionary context signal is exposed to the model during pretraining. At the same time, the later-stage curves partially reconverge, which helps explain why MLM differences remain relatively modest compared with LRCP: MLM is sensitive enough to reveal sampler-induced distributional effects, but not discriminative enough to fully resolve the downstream ranking. Taken together, these observations strengthen the interpretation that MSA diversity and query identity are not incidental byproducts of subsampling, but operative factors that influence both representation learning and downstream behavior.

\section{Experiment Details}
\subsection{Protein Language Model Training Details}
\label{si:mlm}
Following the BERT-style~\citep{bert} masking scheme, 15\% of eligible tokens are selected for prediction; among them, 80\% are replaced by the mask token, 10\% by a random amino acid token, and the remaining 10\% are left unchanged. As the MSA-based PLM backbone for our experiments, we use MSA Transformer~\citep{rao2021msa} with the standard 12-layer, 768-dimensional architecture. All the experiment settings above are consistent with the MSA Transformer paper or BERT masking convention. 

We use the 260K training set of Uniclust30 MSAs from OpenProteinSet~\citep{ahdritz2023openproteinsettrainingdatastructural} for pretraining and a 10K held-out set for testing. We preprocess the raw MSAs offline using the specified subsampling strategy, and then train the PLM on the resulting presampled dataset. Note that test set is random-subsampled as well. We evaluate PLMs pretrained with six different MSA subsampling strategies---random, divMax, HHfilter, divMin, K-medians and \ap{}---for the MLM testing in this section and the downstream Long Range Contact Predition task in Section~\ref{exp:lrcp}. We reveal more information about the baseline subsamplers, model pretraining, dataset and metrics in Appendix~\ref{si:subsampler} and~\ref{si:pretrain}.

\subsection{K-medians as a Baseline}
\label{si:kmedians}
At the first beginning, we wanted a representative clustering-based comparator beyond purely greedy samplers, but AF-cluster was not an ideal choice for this purpose: it was originally designed for multiple conformation prediction and not for fixed-cardinality MSA subsampling. Hence we alternatively choose K-medians primarily as a simple clustering baseline. K-medians produces a fixed number of clusters and allows us to test whether a simple cluster strategy is already sufficient to explain the observed phenomenon. Moreover, K-medians is also a baseline for Affinity Propagation in the paper~\citep{frey2007clustering}.

K-medians-based subsampling can also be viewed as solving an optimization problem, where the objective is to minimize the within-cluster distance between MSA rows and their assigned representatives. K-medians-style subsampling can be viewed as optimizing a fixed-cardinality representative-selection objective:
\begin{equation}
\min_{b\in\{0,1\}^{M}}
\sum_{i=0}^{M-1}
\min_{j:\,b_j=1}
d_{\mathrm{MSA}}(x_i,x_j),
\qquad
\mathrm{s.t.}\quad
\sum_{j=0}^{M-1} b_j=N,\quad b_0=1,
\end{equation}
where $b_j=1$ indicates that row $j$ is selected as a representative and $d_{\mathrm{MSA}}$ denotes the gap-aware Hamming distance. However, it is typically optimized by Lloyd-style alternating minimization—iterating between assignment and center-update steps—rather than by probabilistic message passing over a factor graph, and is generally not interpreted as reasoning process in the domain of machine learning. As a result, while it provides a reasonable clustering-based baseline, it offers limited flexibility for explicitly modeling evolution-aware preferences such as the query identity of the selected cluster representatives.

Our implementation adapts standard Lloyd-style K-medians to categorical protein-sequence data under gap-aware Hamming distance. We fix the number of clusters to the target row budget \(N=\texttt{max\_seqs}\), initialize centers with a K-medians++ scheme while hard-pinning the query row as one center, alternate between nearest-center assignment and center recomputation by column-wise non-gap mode. Eventually, K-medians produces $K$ clusters and selects real sequence closest to the cluster's column-wise non-gap mode center for each cluster. In this sense, K-medians serves as a lightweight clustering baseline and simply utilized to provide more insights and reveal the hump observation.
\subsection{Baseline Subsampling Strategies}
\label{si:subsampler}
For the experiment of PLM pretraining and long range contact predictions, we compare with four most common subsampling strategies and one additional experiment baseline K-medians. The motivation and implementation of K-medians baseline is elaborated in~\ref{si:kmedians} . For other baseline subsamplers, \texttt{random} uniformly samples non-query rows while always preserving the query, providing the simplest budget-compliant reference. \texttt{divMin} greedily selects rows with minimal average Hamming distance to the current subset, thereby concentrating the MSA around highly redundant local neighborhoods. \texttt{divMax}, in contrast, greedily favors rows with large average Hamming distance to the current subset, explicitly promoting global diversity in the retained alignment. \texttt{HHfilter} applies sequence-identity-based redundancy filtering, removing near-duplicate rows while preserving a more coverage-oriented set of homologs.

\subsection{Protein Language Model Pretraining}
\label{si:pretrain}

\paragraph{Pretraining.} We use a per-MSA token budget of \(B=16384\) and a maximum sequence length of 1024. We adopted dropout of 0.1, AdamW with learning rate \(10^{-4}\), weight decay 0.01, \(\beta=(0.9,0.98)\), 16k warmup steps followed by cosine decay over 100k total steps, and bf16 mixed-precision training. For AP-based subsampling, we empirically adopt \(\alpha=1\) and \(\beta=0\) as a balanced operating configuration. We train for 150{,}000 optimization steps, using gradient accumulation over 8 micro-batches per step. All our pretraining experiments are run on NVIDIA A800.

\paragraph{Dataset.}  For pretraining, we use the 260K-MSA training split of OpenProteinSet derived from Uniclust30, and evaluate on a held-out set of 10K MSAs. OpenProteinSet~\citep{ahdritz2023openproteinsettrainingdatastructural} is a large-scale, openly available protein dataset comprising over 16 million MSAs, associated structural homologs retrieved from the Protein Data Bank, and AlphaFold2-predicted structures~\citep{jumper2021highly}. Released to address the prohibitive computational cost of generating MSAs, it has been successfully used to retrain AlphaFold2 and supports a wide range of machine learning applications in protein structure prediction, functional analysis, and multimodal modeling. Uniclust30 is a clustered protein-sequence database in which sequences are grouped at 30\% pairwise identity, providing diverse yet non-redundant multiple-sequence alignments for large-scale protein modeling.

\paragraph{Metrics.} For the MLM task, we report three standard token-level evaluation metrics on the held-out test set: cross-entropy loss, perplexity (PPL), and denoising accuracy. The loss and PPL quantify how well the model predicts masked residues overall, with lower values indicating better performance, while denoising accuracy measures the fraction of masked positions whose original amino acids are recovered exactly. Together, these metrics provide complementary views of probabilistic calibration and discrete recovery quality.

\subsection{Long Range Contact Prediction Settings}
\label{si:lrcp}

\paragraph{Sparse Regression Head.} The prediction head consists of a sparse $L_1$-regularized logistic regression model. The optimization objective is formulated as:$$\min_{w} \sum_{i} \text{BCE}(y_i, \hat{y}_i) + \lambda \|w\|_1$$where $\text{BCE}$ denotes the binary cross-entropy loss and $\lambda = 0.15$ is the regularization strength. This formulation yields a sparse solution where only a subset of the attention heads are active. We specifically employ a coordinate-descent based solver to strictly enforce the $L_1$ penalty, as standard momentum-based optimizers (e.g., Adam) can interfere with the sparsity of the final solution.

\paragraph{Feature Representation.} Predictions are derived exclusively from the row-attention maps of the frozen MSA Transformer (12 layers, 12 heads). Stacking the attention matrices across all layers and heads yields a 144-dimensional feature vector for each residue pair. Before being fed into the regression head, these attention maps are symmetrized and refined using Average Product Correction (APC). Features are processed in high precision to ensure numerical stability during optimization.

\paragraph{Dataset} The RCSB Protein Data Bank (RCSB PDB) is a public repository of experimentally resolved three-dimensional protein structures, providing atomic coordinates from techniques such as X-ray crystallography, NMR, and cryo-EM. In this work, we use a small set of 20 RCSB structures as supervision for fitting the lightweight sparse regression head in unsupervised contact prediction, where the native structures supply residue-residue contact maps while the pretrained protein language model itself remains frozen.

CASP15, held from May to August 2022, is a community-wide blind benchmark for evaluating biomolecular structure prediction methods against experimental ground truth. It comprises 127 targets spanning single protein domains, multimeric assemblies, RNA structures, protein–ligand complexes, and conformational ensembles, with over 53,000 models submitted by approximately 100 research groups worldwide.

\paragraph{Test Set Settings and Metrics} Following standard protein contact-prediction evaluation, we report Top-\(L\), Top-\(L/2\), and Top-\(L/5\) precision, where \(L\) denotes the length of the target protein (or evaluated domain). For each target, all candidate residue pairs are ranked by predicted contact probability, and precision is computed over the top \(K\) predictions for \(K \in \{L, L/2, L/5\}\). Here, precision is defined as \(\mathrm{TP}/(\mathrm{TP}+\mathrm{FP})\), i.e., the fraction of the top-ranked predicted residue pairs that are true contacts in the native structure. We restrict evaluation to long-range pairs satisfying \(|i-j| \geq 24\), so that the metric emphasizes nonlocal co-evolutionary structure rather than trivial local backbone geometry. Performance is evaluated on 45 CASP15 monomer domains. Following MSA Transformer~\citep{rao2021msa}, which showed that divMax MSA subsampling better reflects a PLM's long-range contact prediction capability, we evaluate on CASP15 using divMax subsampling at test time.

\paragraph{Informative Subset} The performance analysis is conducted on the Informative Subset of 36 CASP15 domains, excluding 9 targets where structural reconstruction is fundamentally precluded by data constraints. These excluded domains are primarily characterized by extreme sequence sparsity, where the $N_{\text{eff}}$ is too low to distinguish meaningful co-evolutionary signals from stochastic noise, rendering MSA-based inference physically groundless. Furthermore, several targets represent "orphan folds" or de novo designed proteins that lack discernible evolutionary lineages. By focusing on this subset, we ensure that the evaluation accurately reflects subsamplers' capacity to interpret biophysically well-defined evolutionary information rather than introducing bias from data-deficient targets.

\subsection{Protein Multiple Conformational Predictions}
\label{si:pmcp_box_stats}

\paragraph{Dataset and metrics.} In particular, we evaluate on three well-characterized fold-switching proteins: KaiB, RfaH, and MAD2, possessing at least two experimentally deposited conformational states. For each prediction, we report RMSD and pLDDT relative to these references. To assess landscape coverage, we report the empirical distribution of predictions across states.

\paragraph{Baselines.}
We evaluate \textsc{\ap{}} against two representative baselines using an identical frozen AlphaFold2 backbone to isolate the specific impact of MSA selection on conformational recovery. The first baseline is \textbf{AF-Cluster}, an established MSA-subsampling method that identifies alternative states by clustering homologs within a sequence-embedding space. The second baseline is \textbf{Random} sampling, which serves as a controlled reference by selecting MSA rows uniformly at random. To ensure a fair comparison, the Random baseline utilizes the same depth budget across 96 seeds. AF-Cluster is executed using its official implementation and default clustering configurations.


\begin{table}[h]
\centering
\small
\setlength{\tabcolsep}{4pt}
\renewcommand{\arraystretch}{1.00}
\caption{Per-cluster recovery statistics on PMCP. $n$ = number of confident top-$1$ predictions inside the cluster box; $\varnothing$ = empty cluster. Bold entries highlight the within-row optimum (lower RMSD or higher pLDDT).}
\label{tab:pmcp_box_stats}
\begin{tabular}{l l l r r r r r}
\toprule
Protein & State & Method & $n$ & avg RMSD (\AA) & avg pLDDT & best RMSD (\AA) & best pLDDT \\
\midrule
KaiB & Ground     & \ap{}     & 12   & \textbf{1.74} & \textbf{80.0} & 1.57 & \textbf{84.9} \\
KaiB & Ground     & AF-Cluster  & 27   & 2.15 & 75.2 & \textbf{1.47} & 80.5 \\
KaiB & Ground     & Random     & 0    & $\varnothing$ & $\varnothing$ & --   & --   \\
\midrule
KaiB & FS         & \ap{}     & 56   & \textbf{2.20} & \textbf{88.8} & \textbf{1.51} & 91.2 \\
KaiB & FS         & AF-Cluster  & 1008 & 2.35 & 84.7 & 1.59 & \textbf{91.8} \\
KaiB & FS         & Random     & 96   & 2.33 & 88.7 & 2.17 & 91.7 \\
\midrule
Mad2 & Closed     & \ap{}     & 50   & 4.17 & 90.0 & \textbf{2.86} & 91.7 \\
Mad2 & Closed     & AF-Cluster  & 18   & 4.06 & 86.9 & 2.99 & 91.6 \\
Mad2 & Closed     & Random     & 96   & \textbf{3.94} & \textbf{92.3} & 3.70 & \textbf{93.1} \\
\midrule
Mad2 & Open       & \ap{}$^{\ddagger}$  & 6 & 8.54 & \textbf{69.9} & 6.67 & \textbf{78.8} \\
Mad2 & Open       & AF-Cluster$^{\ddagger}$ & 6 & \textbf{7.42} & 58.5 & \textbf{5.32} & 64.0 \\
Mad2 & Open       & Random     & 0    & $\varnothing$ & $\varnothing$ & --   & --   \\
\midrule
RfaH & Autoinhib. & \ap{}$^{\dagger}$    & 22 & \textbf{3.55} & 73.3 & \textbf{2.15} & 77.9 \\
RfaH & Autoinhib. & AF-Cluster             & 4  & 3.57 & 71.3 & 2.57 & 73.4 \\
RfaH & Autoinhib. & Random$^{\dagger}$    & 16 & 4.86 & \textbf{74.1} & 4.54 & \textbf{78.5} \\
\midrule
RfaH & Active     & \ap{}     & 36 & 4.26 & 83.8 & 3.53 & \textbf{92.7} \\
RfaH & Active     & AF-Cluster  & 6  & \textbf{3.82} & 80.9 & \textbf{2.91} & 84.9 \\
RfaH & Active     & Random     & 27 & 3.98 & \textbf{88.6} & 3.20 & 92.9 \\
\bottomrule
\end{tabular}
\end{table}

\section{Additional Experiment Results and Visualizations}
\label{si:lrcp_more}
\subsection{Additional Top-L/2 and Top-L/5 Results on divMax-Sampled CASP15}
As a complement to the main-text Top-L comparison, we provide the corresponding Top-L/2 and Top-L/5 results here for the same six checkpoints under divMax-sampled CASP15 evaluation.
\begin{table}[h]
\centering
\small
\setlength{\tabcolsep}{6pt}
\renewcommand{\arraystretch}{1.05}
\caption{Top-L/2 long-range contact precision (\%) on divMax-sampled CASP15 MSAs. Columns denote training-time samplers (checkpoints).}
\label{tab:divmax-casp15-topl2}
\begin{tabular}{lcccccc}
\toprule
Setting & random & divMin & divMax & kmedians & HHfilter & AP \\
\midrule
Full CASP15 ($n=45$)        & 41.8 & 22.9 & 37.3 & 35.2 & 29.6 & \textbf{47.3} \\
Informative subset ($n=36$) & 51.5 & 28.0 & 46.0 & 43.5 & 36.3 & \textbf{58.2} \\
\bottomrule
\end{tabular}
\end{table}

\begin{table}[h]
\centering
\small
\setlength{\tabcolsep}{6pt}
\renewcommand{\arraystretch}{1.05}
\caption{Top-L/5 long-range contact precision (\%) on divMax-sampled CASP15 MSAs. Columns denote training-time samplers (checkpoints).}
\label{tab:divmax-casp15-topl5}
\begin{tabular}{lcccccc}
\toprule
Setting & random & divMin & divMax & kmedians & HHfilter & AP \\
\midrule
Full CASP15 ($n=45$)        & 53.0 & 32.7 & 50.0 & 46.7 & 39.1 & \textbf{58.0} \\
Informative subset ($n=36$) & 65.0 & 40.0 & 61.9 & 57.4 & 47.9 & \textbf{71.4} \\
\bottomrule
\end{tabular}
\end{table}

\subsection{Additional LRCP Precision Visualizations}
\begin{figure}[htbp] 
    \centering 
    \includegraphics[width=0.75\textwidth]{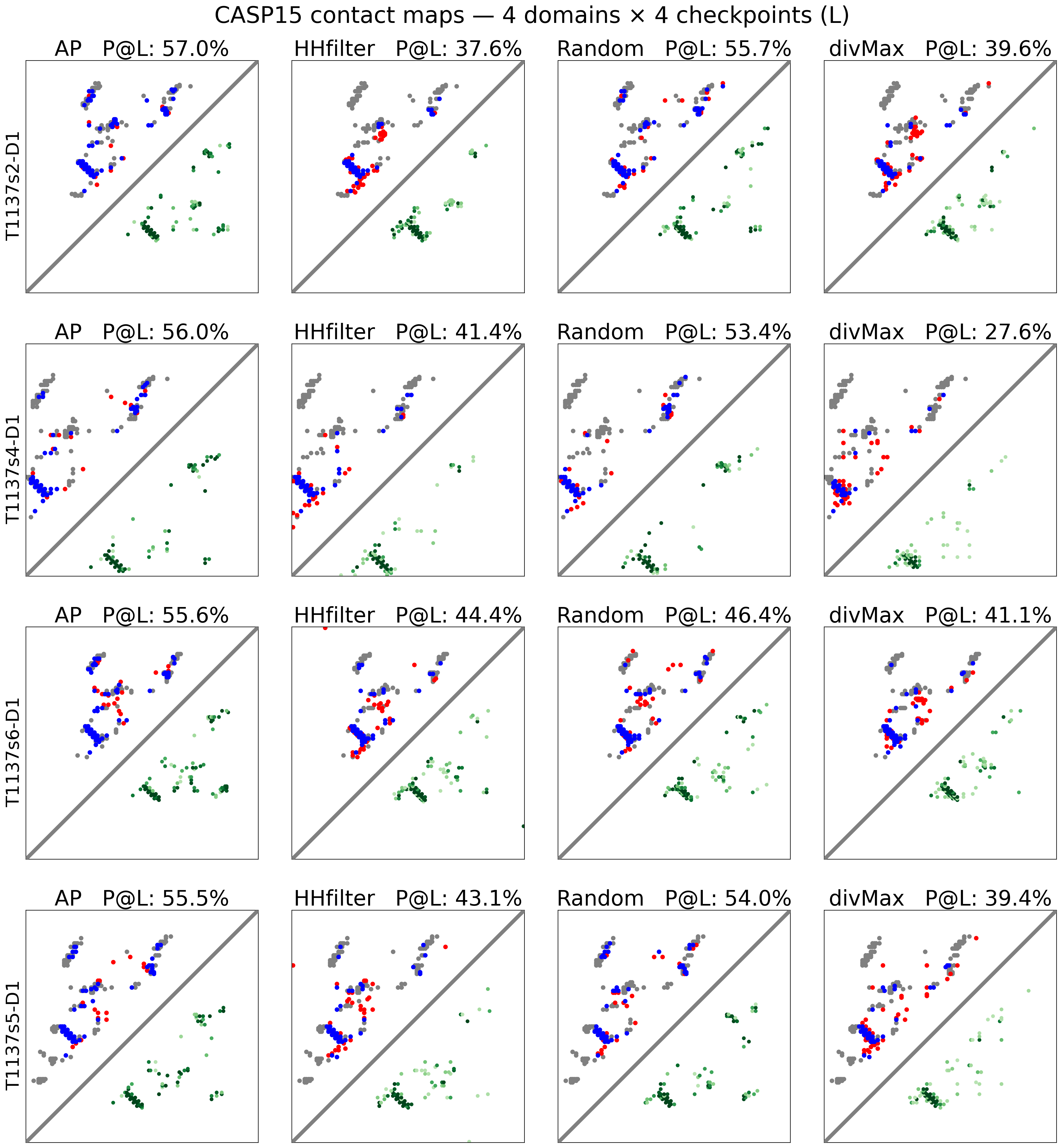} 
    \caption{Contact prediction under different MSA subsampling strategies on CASP15 targets. We compare AP, HHfilter, Random and divMax on four additional domains, reporting Top-L contact precision P@L above each contact map. Raw contact probabilities are represented by green points below the diagonal, while Top-L predicted contacts are shown above the diagonal; blue, red, and grey points denote true positives, false positives, and ground-truth contacts, respectively.}
    \label{fig:si_lrcp}
\end{figure}

Here we present four additional long range contact prediction precision visualization cases in Figure~\ref{fig:si_lrcp}.

\subsection{Hump}
we plot the downstream long range contact precision against both sequence diversity and query identity to visualize this trade-off. Presented in Figure~\ref{fig:hump}, we observe a hump-shaped correlation linking contact precision to both axes, uncovering a comfort-zone inaccessible as for traditional single-axis samplers. Consequently, the \ap{} achieves the highest downstream performance by elegantly resolving the trade-off between preserving query identity and maximizing evolutionary variance through reasoning over the optimization. More hump details and more evidence about the reasonability for optimization formulation can be referred to Appendix~\ref{si:hump}.

\begin{figure}[htbp] 
    \centering 
    \includegraphics[width=0.9\textwidth]{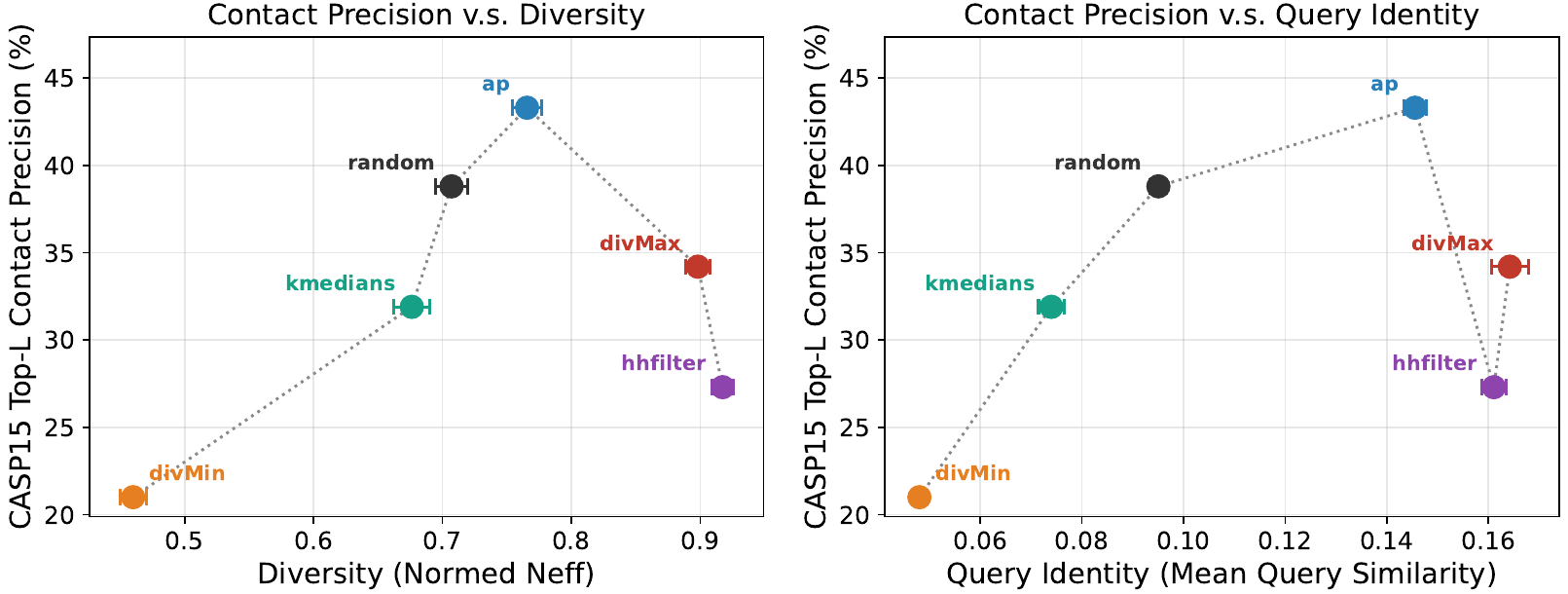} 
    \caption{Training-MSA Diversity / Query Identity v.s. Informative CASP15 Top-L Contact Precision}
    \label{fig:hump}
\end{figure}

\subsection{Hump Result on Full CASP15}
\begin{figure}[htbp] 
    \centering 
    \includegraphics[width=0.9\textwidth]{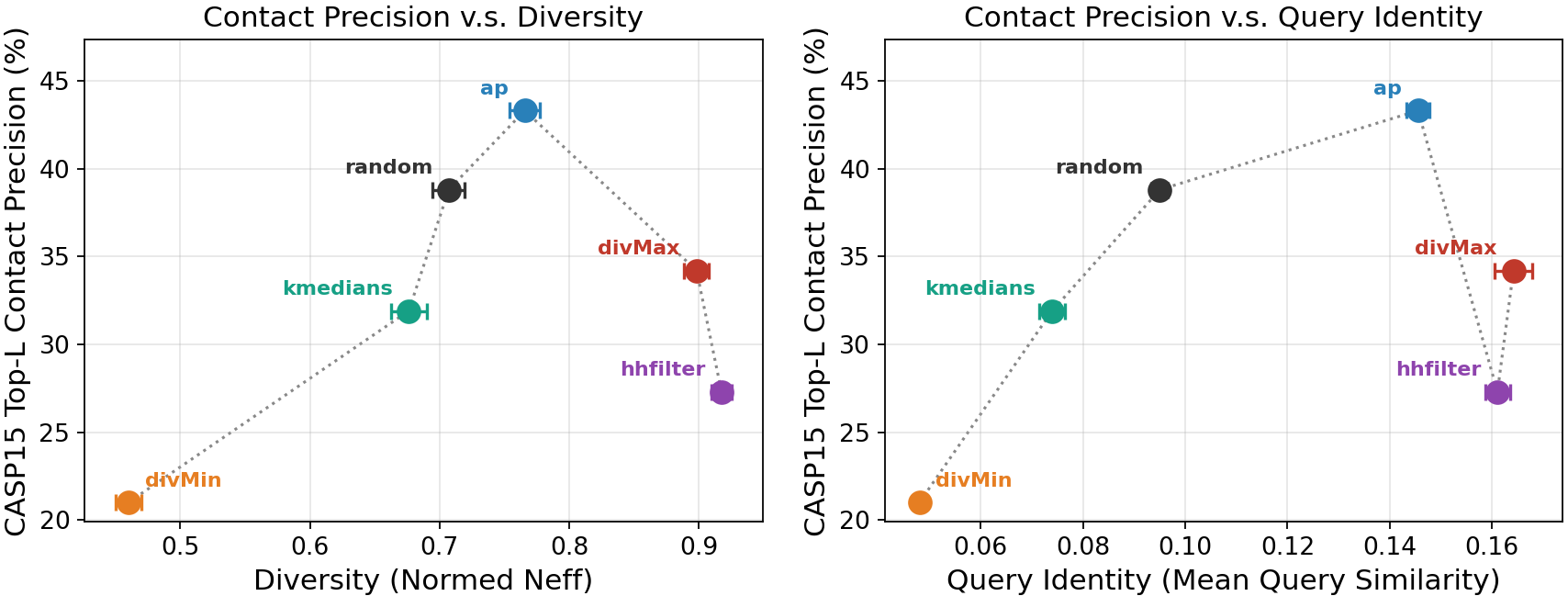} 
    \caption{Training-MSA Diversity / Query Identity v.s. full CASP15 Top-L Contact Precision}
    \label{fig:si_hump}
\end{figure}

Hereby we present the full result figure on CASP15 as a complement. The hump can be observed as well in Figure~\ref{fig:si_hump}.

\subsection{$\alpha$ Control Knob}
\label{si:alpha-result}
Hereby we report the remaining five tables on $\alpha$ control knob experiment in long range contact prediction. These tables preserve the same qualitative conclusion as in the main text: the effect of $\alpha$ is clearest under Top-L/5 budget, while under the looser Top-L and Top-L/2 the trend weakens but still exists. Visualization about the col mean and alpha is presented here as well. 

\begin{figure}[htbp]
  \centering
  \includegraphics[width=\linewidth]{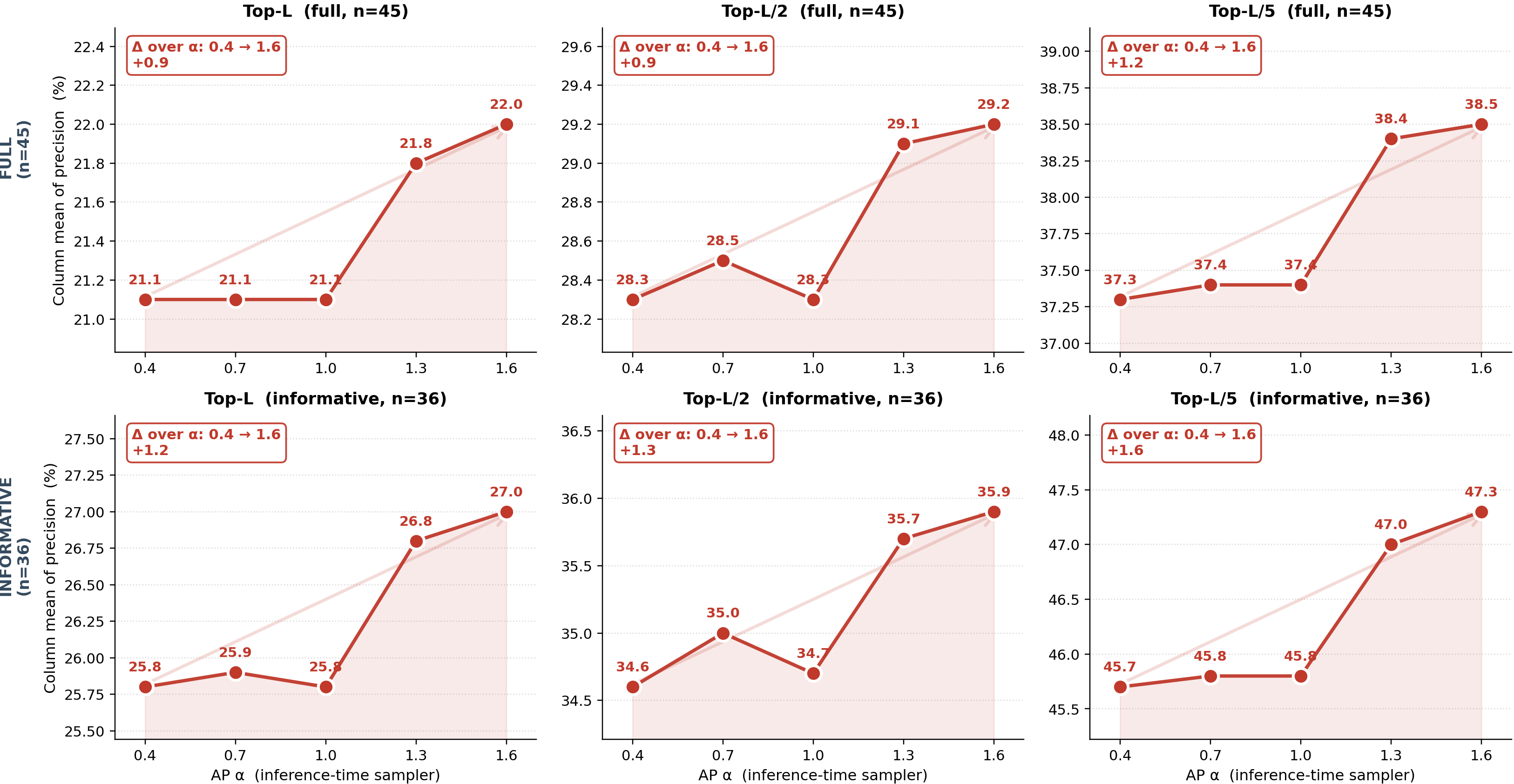}
  \caption{Column-mean precision v.s. AP $\alpha$ knob under all settings.}
\end{figure}

\begin{table}[htbp]
\centering
\small
\setlength{\tabcolsep}{4pt}
\caption{Top-L CASP15 long range contact prediction results on the full set ($n=45$).}
\label{tab:topl_full_alpha}
\resizebox{\linewidth}{!}{
\begin{tabular}{lcccccc}
\toprule
ckpt & AP ($\alpha=0.4$) & AP ($\alpha=0.7$) & AP ($\alpha=1.0$) & AP ($\alpha=1.3$) & AP ($\alpha=1.6$) & divMax \\
\midrule
random   & 25.4 & 25.3 & 25.5 & 26.3 & 25.8 & 27.6 \\
divMax   & 22.9 & 23.2 & 21.6 & 23.3 & 23.1 & 27.3 \\
HHfilter & 16.6 & 16.3 & 16.3 & 17.2 & 17.2 & 20.3 \\
divMin   & 11.4 & 11.6 & 11.7 & 12.2 & 12.7 & 12.2 \\
ap       & 29.0 & 29.2 & 29.2 & 29.2 & 29.4 & 31.7 \\
kmedians & 21.0 & 21.2 & 21.9 & 22.6 & 23.6 & 19.3 \\
\midrule
col mean & 21.1 & 21.1 & 21.1 & 21.8 & 22.0 & 23.1 \\
\bottomrule
\end{tabular}
}
\end{table}

\begin{table}[htbp]
\centering
\small
\setlength{\tabcolsep}{4pt}
\caption{Top-L CASP15 long range contact prediction results on the 36-domain subset obtained after excluding 9 universal-failure domains.}
\label{tab:topl_informative_alpha}
\resizebox{\linewidth}{!}{
\begin{tabular}{lcccccc}
\toprule
ckpt & AP ($\alpha=0.4$) & AP ($\alpha=0.7$) & AP ($\alpha=1.0$) & AP ($\alpha=1.3$) & AP ($\alpha=1.6$) & divMax \\
\midrule
random   & 31.2 & 31.2 & 31.4 & 32.3 & 31.8 & 34.1 \\
divMax   & 28.2 & 28.7 & 26.7 & 28.8 & 28.5 & 33.7 \\
HHfilter & 20.3 & 19.9 & 19.9 & 21.1 & 21.1 & 25.0 \\
divMin   & 13.7 & 14.0 & 14.1 & 14.7 & 15.4 & 14.7 \\
ap       & 35.7 & 35.8 & 35.9 & 35.8 & 36.1 & 39.0 \\
kmedians & 25.8 & 26.0 & 27.0 & 27.8 & 29.1 & 23.6 \\
\midrule
col mean & 25.8 & 25.9 & 25.8 & 26.8 & 27.0 & 28.4 \\
\bottomrule
\end{tabular}
}
\end{table}

\begin{table}[htbp]
\centering
\small
\setlength{\tabcolsep}{4pt}
\caption{Top-L/2 CASP15 long range contact prediction results on the full set ($n=45$).}
\label{tab:topl2_full_alpha}
\resizebox{\linewidth}{!}{
\begin{tabular}{lcccccc}
\toprule
ckpt & AP ($\alpha=0.4$) & AP ($\alpha=0.7$) & AP ($\alpha=1.0$) & AP ($\alpha=1.3$) & AP ($\alpha=1.6$) & divMax \\
\midrule
random   & 34.1 & 33.6 & 34.0 & 34.8 & 34.2 & 37.1 \\
divMax   & 30.7 & 31.4 & 29.2 & 30.6 & 30.7 & 36.3 \\
HHfilter & 21.9 & 22.0 & 22.0 & 22.9 & 23.8 & 27.7 \\
divMin   & 15.1 & 15.4 & 15.7 & 16.5 & 16.4 & 16.4 \\
ap       & 39.2 & 39.5 & 39.0 & 39.0 & 38.9 & 42.3 \\
kmedians & 28.6 & 29.2 & 29.8 & 30.5 & 31.2 & 26.4 \\
\midrule
col mean & 28.3 & 28.5 & 28.3 & 29.1 & 29.2 & 31.1 \\
\bottomrule
\end{tabular}
}
\end{table}

\begin{table}[htbp]
\centering
\small
\setlength{\tabcolsep}{4pt}
\caption{Top-L/2 CASP15 long range contact prediction results on the 36-domain subset obtained after excluding 9 universal-failure domains.}
\label{tab:topl2_informative_alpha}
\resizebox{\linewidth}{!}{
\begin{tabular}{lcccccc}
\toprule
ckpt & AP ($\alpha=0.4$) & AP ($\alpha=0.7$) & AP ($\alpha=1.0$) & AP ($\alpha=1.3$) & AP ($\alpha=1.6$) & divMax \\
\midrule
random   & 41.9 & 41.4 & 42.0 & 42.9 & 42.2 & 45.8 \\
divMax   & 37.9 & 38.7 & 36.0 & 37.8 & 37.8 & 45.0 \\
HHfilter & 26.8 & 26.9 & 26.9 & 28.2 & 29.0 & 34.1 \\
divMin   & 18.2 & 18.7 & 18.9 & 19.9 & 19.9 & 19.9 \\
ap       & 48.0 & 48.5 & 47.7 & 47.7 & 47.9 & 52.0 \\
kmedians & 35.1 & 36.0 & 36.8 & 37.6 & 38.4 & 32.5 \\
\midrule
col mean & 34.6 & 35.0 & 34.7 & 35.7 & 35.9 & 38.2 \\
\bottomrule
\end{tabular}
}
\end{table}

\begin{table}[h]
\centering
\small
\setlength{\tabcolsep}{4pt}
\caption{Top-L/5 CASP15 long range contact prediction results on the full set ($n=45$).}
\label{tab:topl5_full_alpha}
\resizebox{\linewidth}{!}{
\begin{tabular}{lcccccc}
\toprule
ckpt & AP ($\alpha=0.4$) & AP ($\alpha=0.7$) & AP ($\alpha=1.0$) & AP ($\alpha=1.3$) & AP ($\alpha=1.6$) & divMax \\
\midrule
random   & 43.8 & 43.5 & 43.6 & 44.4 & 44.7 & 48.0 \\
divMax   & 41.2 & 42.1 & 40.1 & 41.7 & 42.3 & 48.9 \\
HHfilter & 30.6 & 30.5 & 31.3 & 31.8 & 32.3 & 36.4 \\
divMin   & 21.1 & 20.2 & 21.2 & 22.1 & 21.9 & 23.3 \\
ap       & 49.6 & 50.3 & 49.4 & 49.5 & 49.1 & 52.6 \\
kmedians & 37.3 & 37.9 & 38.8 & 40.6 & 40.7 & 35.7 \\
\midrule
col mean & 37.3 & 37.4 & 37.4 & 38.4 & 38.5 & 40.8 \\
\bottomrule
\end{tabular}
}
\end{table}

\begin{figure}
    \centering
    \includegraphics[width=1\linewidth]{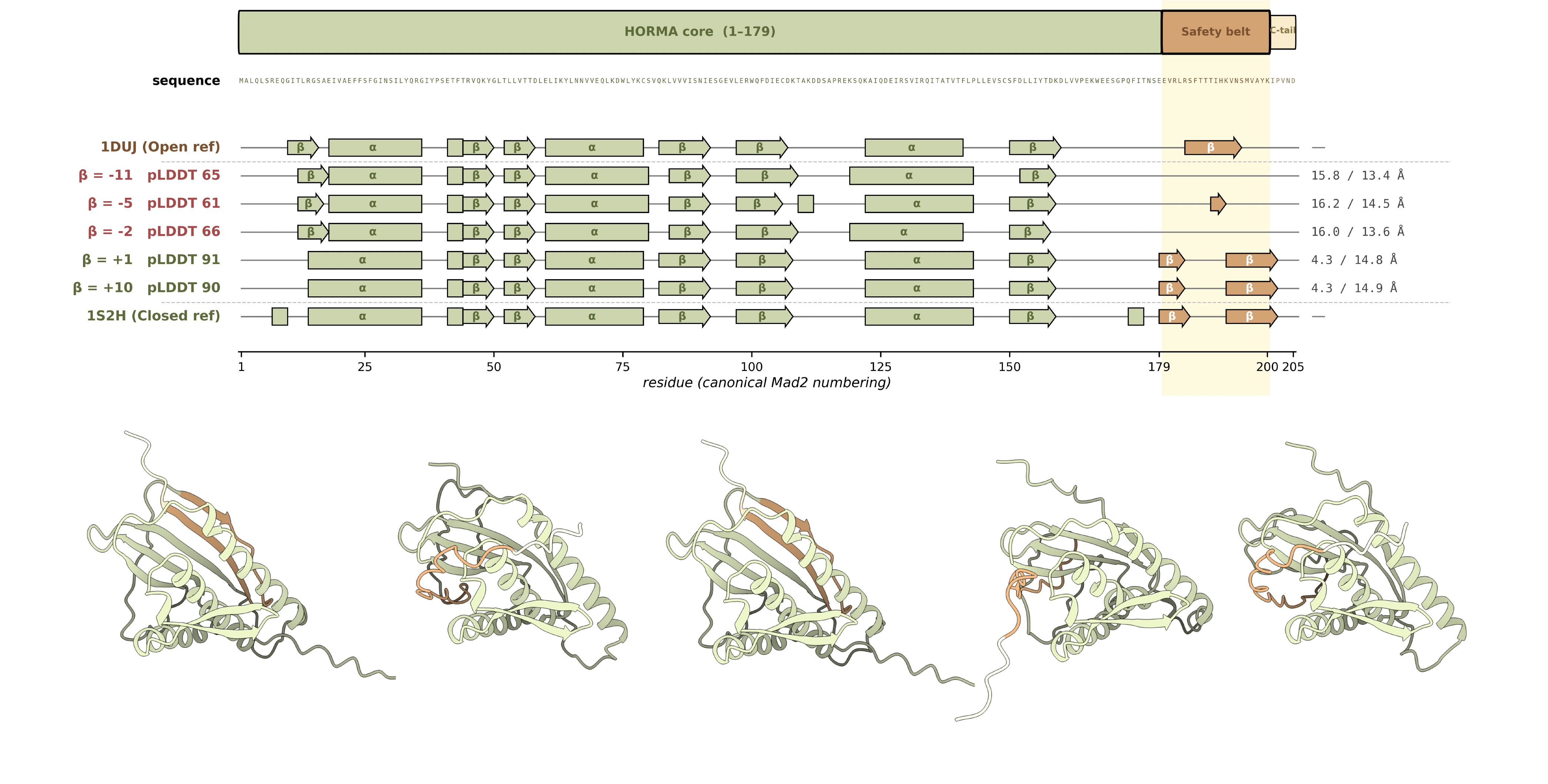}
    \caption{Structural transition of KaiB controled by the $\beta$ knob.}
    \label{fig:MAD_SI}
\end{figure}

\begin{figure}
    \centering
    \includegraphics[width=1\linewidth]{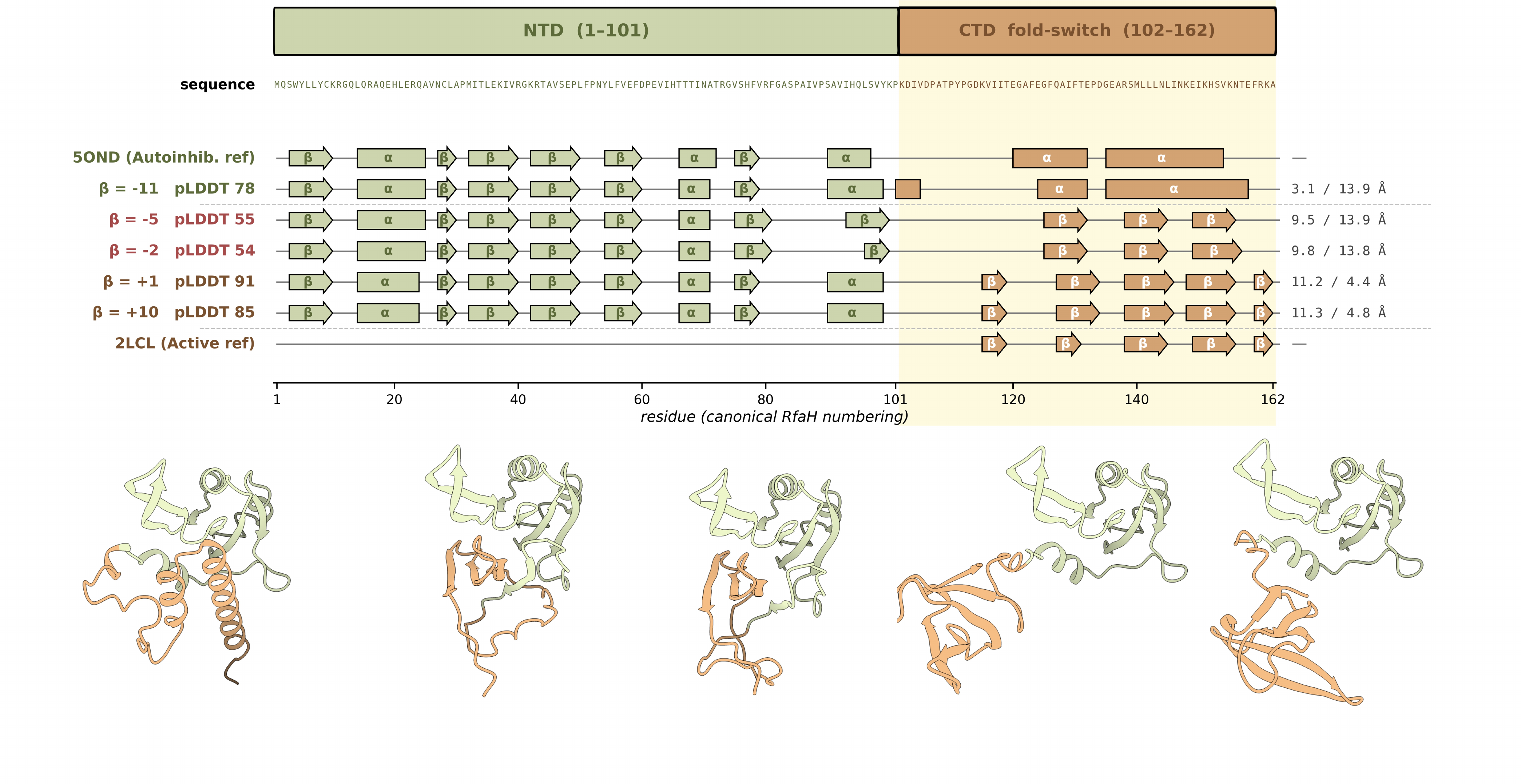}
    \caption{Structural transition of RafH controled by the $\beta$knob.}
    \label{fig:RafH_SI}
\end{figure}


\end{document}